\documentclass[lettersize,journal]{IEEEtran}

\usepackage{bm}
\usepackage{verbatim}
\usepackage{amsmath,amsfonts}
\usepackage{cases}
\usepackage{booktabs} 
\usepackage{graphicx}
\usepackage{multirow}
\usepackage{subfigure}
\usepackage{array}
\usepackage{textcomp}
\usepackage{stfloats}
\usepackage{url}
\usepackage{cite}
\usepackage{amsthm,amsmath,amssymb}
\usepackage{mathrsfs}
\usepackage{bbding}
\usepackage{xcolor}

\usepackage{booktabs}
\usepackage{threeparttable}

\newtheorem{lemma}{Lemma}
\usepackage{algorithm}
\usepackage{algorithmic}
\usepackage{hyperref}
\hypersetup{hypertex=true,
            colorlinks=true,
            linkcolor=red, 
            anchorcolor=blue, 
            citecolor=blue}

\begin{document}

\title{Efficient Multi-view Clustering via Unified and Discrete Bipartite Graph Learning}

\author{Si-Guo Fang,
	Dong Huang,
    Xiao-Sha Cai,
	Chang-Dong Wang,
	Chaobo He,
	and~Yong Tang
	\IEEEcompsocitemizethanks{\IEEEcompsocthanksitem S.-G. Fang and D. Huang are with the College of Mathematics and Informatics, South China Agricultural University, Guangzhou, China, and also with Key Laboratory of Smart Agricultural Technology in Tropical South China, Ministry of Agriculture and Rural Affairs, China. \protect\\
		E-mail: siguofang@hotmail.com, huangdonghere@gmail.com.
		\IEEEcompsocthanksitem X.-S. Cai and C.-D. Wang are with the School of Computer Science and Engineering,
		Sun Yat-sen University, Guangzhou, China, and also with Guangdong Key Laboratory of Information Security Technology, Guangzhou, China.
		E-mail: xiaoshacai@hotmail.com, changdongwang@hotmail.com.
		\IEEEcompsocthanksitem C. He and Y. Tang are with the School of Computer Science, South China Normal University, Guangzhou, China.
		E-mail: hechaobo@foxmail.com, ytang@m.scnu.edu.cn.
}
	}

\markboth{}
{Shell \MakeLowercase{\textit{et al.}}: Bare Demo of IEEEtran.cls for Computer Society Journals}

\IEEEtitleabstractindextext{
	\begin{abstract}
		Although previous graph-based multi-view clustering algorithms have gained significant progress, most of them are still faced with three limitations. First, they often suffer from high computational complexity, which restricts their applications in large-scale scenarios. Second, they usually perform graph learning either at the single-view level or at the view-consensus level, but often neglect the possibility of the joint learning of single-view and consensus graphs. Third, many of them rely on the $k$-means for discretization of the spectral embeddings, which lack the ability to directly learn the graph with discrete cluster structure.  In light of this, this paper presents an efficient multi-view clustering approach via \textit{\textbf{u}}nified and \textit{\textbf{d}}iscrete \textit{\textbf{b}}ipartite \textit{\textbf{g}}raph \textit{\textbf{l}}earning (UDBGL). Specifically, the anchor-based subspace learning is incorporated to learn the view-specific bipartite graphs from multiple views, upon which the bipartite graph fusion is leveraged to learn a view-consensus bipartite graph with adaptive weight learning. Further, the Laplacian rank constraint is imposed to ensure that the fused bipartite graph has discrete cluster structures (with a specific number of connected components). By simultaneously formulating the view-specific bipartite graph learning, the view-consensus bipartite graph learning, and the discrete cluster structure learning into a unified objective function, an efficient minimization algorithm is then designed to tackle this optimization problem and directly achieve a discrete clustering solution without requiring additional partitioning, which notably has linear time complexity in data size. Experiments on a variety of multi-view datasets demonstrate the robustness and efficiency of our UDBGL approach. The code is available at \url{https://github.com/huangdonghere/UDBGL}.

	\end{abstract}
	
	\begin{IEEEkeywords}
		Data clustering, Multi-view clustering, Large-scale clustering, Bipartite graph learning, Linear time.
\end{IEEEkeywords}}

\maketitle

\IEEEdisplaynontitleabstractindextext

\IEEEpeerreviewmaketitle

\section{Introduction}\label{sec:introduction}

\IEEEPARstart{W}{ith} the continuous development of information technology, an increasing amount of data can be collected from multiple sources (or views) in real-world scenarios, which is often referred to as multi-view data \cite{chao2021survey}. For example, an image can be depicted by different feature descriptors, such as local binary pattern (LBP), histogram of oriented gradient (HOG), and Gabor descriptor. A piece of news can be described in different languages, such Chinese, English, and Spanish. To effectively utilize the multi-view information for the clustering task, many multi-view clustering (MVC) methods have been developed in the literature, among which the graph-based methods  have attracted significant research attention in recent years.

The graph-based MVC methods \cite{SwMC,MCGC,GMC,liangTNNLS,RG-MVC,AMGL,Li2022} typically build (or learn) some graph structure(s) to reflect the sample-wise relationships in multi-view data, and then partition the graph to obtain the clustering result. A general strategy is to formulate the sample-wise relationships into an $n\times n$ adjacent graph, where $n$ is the number of samples. Wang et al.\cite{GMC} utilized a mutual reinforcement technique to learn the single-view graphs and a unified global graph. Liang et al. \cite{liangICDM,liangTNNLS} leveraged both the consistency and inconsistency of multiple single-view graphs, and learned a unified graph via multi-view graph learning. Besides these graph learning based methods \cite{GMC,liangICDM,liangTNNLS,RG-MVC}, another popular sub-category of the graph-based MVC methods are the subspace learning based methods \cite{gao2015multi,zhang2017latent,vidal2011subspace,zhang2020one,tang19_tmm,Kang2020}, which seek to learn a self-representation matrix from low-dimensional subspaces and further utilize the learned self-representation matrix as a global similarity graph for final clustering. Cao et al. \cite{DiMSC} proposed a multi-view subspace clustering method via smoothness and diversity, with the complementarity of multiple representations investigated. Chen et al. \cite{ChenManSheng} enabled the simultaneous learning of the latent embedding space, the global self-representation, and the cluster structures for multi-view subspace learning. Despite the rapid progress, yet these methods mostly suffer from high computational complexity, where the construction of an $n\times n$ adjacent graph typically consumes $\mathcal{O}(n^2)$ time and the spectral partitioning of this graph may even take $\mathcal{O}(n^3)$ time, which restrict their computational feasibility for large datasets.

\begin{figure*}[!th]
\centering
\includegraphics[width=0.97\textwidth]{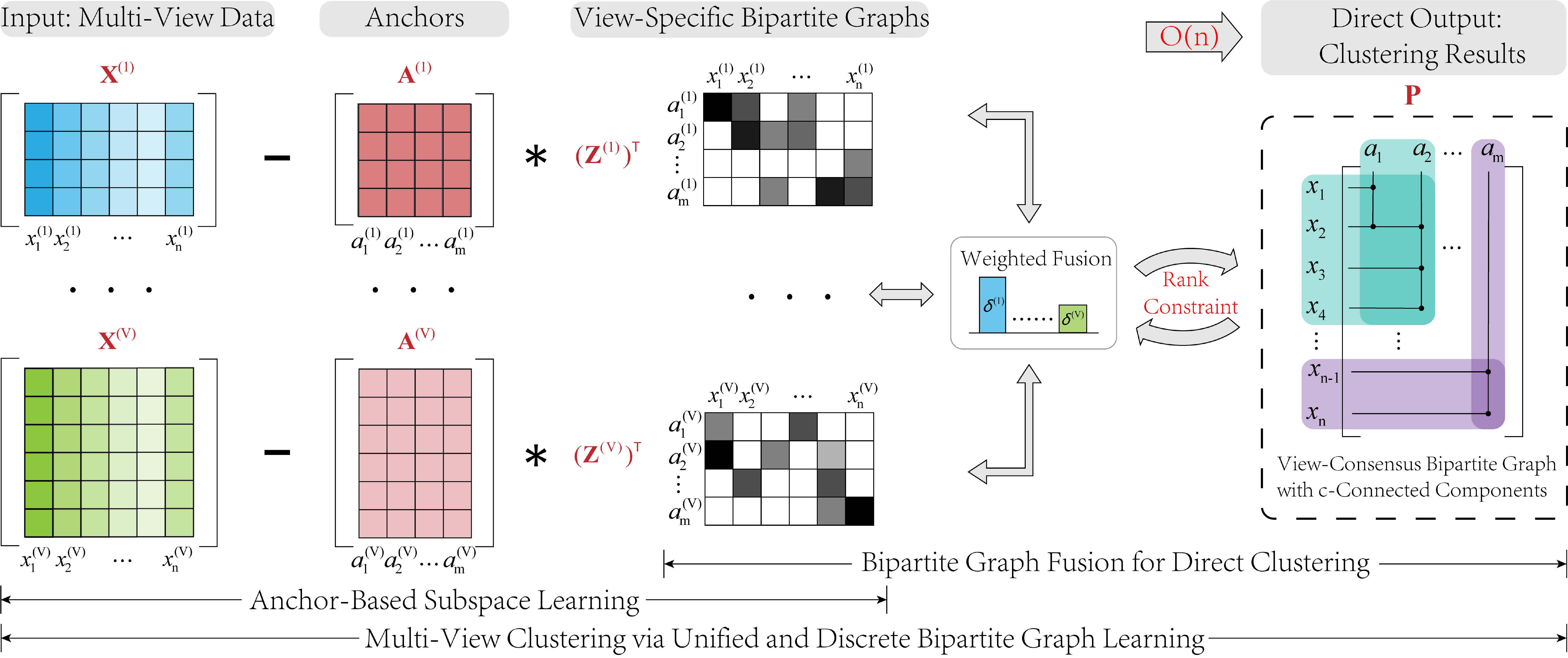}
\caption{Illustration of our UDBGL framework. In the anchor-based subspace learning, multiple view-specific bipartite graphs are learned via the self-expressive loss. In the bipartite graph fusion, UDBGL learns the view-consensus bipartite graph with $c$-connected components and adaptive weights. The view-specific bipartite graph learning (via anchor-based subspace learning) and the view-consensus bipartite graph learning are simultaneously performed and mutually promoted by each other, whose computational complexity is linear to $n$.}
\label{fig:UDBGL}
\end{figure*}

To breakthrough the computational bottleneck, the bipartite graph (i.e., anchor graph) based MVC methods \cite{MVSC,co-clustering-nie2017,guo2019anchors,BIGMC,SFMC,MSGL,nie2021learning,SMVSC,FPMVS-CAG,LMVSC,SDAGF} have recently shown promising capability. Instead of relying on some $n\times n$ graph, the bipartite graph-based methods typically generate a small set of anchors from the original data and then construct an $n\times m$ bipartite graph to represent the data structure \cite{huang2019ultra}, where $m$ is the number of anchors. Specifically, Kang et al. \cite{LMVSC} learned a bipartite graph for each view via the anchor-based subspace learning, and then heuristically concatenated the multiple bipartite graphs into a unified one.
Wang et al. \cite{FPMVS-CAG} learned a unified bipartite graph for all views by means of a set of latent consensus anchors and multiple projection matrices.  However, these bipartite graph-based MVC methods \cite{LMVSC,SMVSC,FPMVS-CAG} either (i) learn a single bipartite graph for all views \cite{SMVSC,FPMVS-CAG} or (ii) learn a bipartite graph for each view and then concatenate these multiple bipartite graphs heuristically \cite{LMVSC}, which lack the ability to simultaneously enforce the view-specific bipartite graph learning and the view-consensus bipartite graph learning.  Furthermore, they mostly regard the bipartite graph learning and its spectral partitioning as two separate phases, yet cannot directly obtain the clustering result by learning a unified bipartite graph with discrete cluster structure.
More recently, Li et al. \cite{SFMC} proposed a scalable multi-view clustering method via bipartite graph fusion. Although it is able to adaptively learn a unified graph with discrete cluster structure by fusing multiple view-specific bipartite graphs, yet these view-specific bipartite graphs are pre-constructed and cannot be adaptively optimized. Despite these recent advances, it is still a challenging problem how to enable the unified learning of the view-specific bipartite graphs and the view-consensus bipartite graph, while ensuring the discrete cluster structure of the learned graph and maintaining the efficiency of the overall framework for large-scale datasets.

To address the above-mentioned problem, this paper presents an efficient multi-view clustering approach via \textbf{u}nified and \textbf{d}iscrete \textbf{b}ipartite \textbf{g}raph \textbf{l}earning (UDBGL).  Particularly, the anchor-based subspace learning is utilized to learn the view-specific bipartite graphs (for multiple views), which are fused into a view-consensus bipartite graph that in turn promotes the view-specific bipartite graph learning. During the fusion, the view weights are adaptively learned to balance the influences of different view-specific bipartite graphs. Furthermore, we impose a Laplacian rank constraint on the unified bipartite graph so as to directly obtain the desired number of connected components, which leads to the discrete cluster structure without requiring additional partitioning. Notably, the view-specific bipartite graph learning, the view-consensus bipartite graph learning, and the discrete cluster structure learning are seamlessly incorporated into a unified objective function, upon which an efficient alternating optimization algorithm is designed to achieve the final clustering with almost linear computational complexity. Experiments are conducted on ten real-world multi-view datasets, which confirm the superiority of our UDBGL approach over the state-of-the-art.

For clarity, the main contributions of this work are summarized as follows:
\begin{itemize}
  \item This paper bridges the gap between the view-specific bipartite graph learning and the view-consensus bipartite graph learning, which can be promoted by each other mutually and adaptively during their learning.
  \item A unified objective function is presented to encompass the view-specific and view-consensus bipartite graph learning and the discrete cluster structure learning, to solve which a \textit{linear-time} optimization algorithm is further designed.
  \item An efficient MVC approach termed UDBGL is proposed. Experiments on multiple multi-view datasets demonstrate the clustering robustness and scalability of our UDBGL approach over the state-of-the-art MVC approaches.
\end{itemize}

The remainder of the paper is arranged as follows. Section~\ref{sec:related_work} reviews the related works. Section~\ref{sec:method} describes the formulation of our UDBGL approach. Section~\ref{sec:opt} presents the optimization algorithm and its theoretical analysis. Section~\ref{sec:experiments} reports the experimental results. Finally, Section~\ref{sec:conclusion} concludes this paper.

\section{Related Work}\label{sec:related_work}

This paper focuses on efficient MVC with unified and discrete bipartite graph learning. Therefore, in this section, our literature review pays attention to three main issues in MVC, namely, how to fuse multiple similarity graphs (from multiple views) into a unified similarity graph, how to build and fuse the similarity graphs with high efficiency (for large datasets), and how to learn the graph with discrete cluster structure without additional partitioning process.

In terms of graph learning (or graph fusion), a variety of MVC methods \cite{MCGC,GMC,liangICDM,liangTNNLS,AMGL} have been proposed, which aim to fuse the similarity relationships among data samples in multiple views into a unified form.
Zhan et al. \cite{MCGC} adopted the rank constraint of the Laplacian matrix and the graph regularization to learn the unified graph from the multiple single-view graphs. Wang et al.\cite{GMC} fused multiple single-view graphs into a unified one, and in turn optimized the single-view graphs through the unified graph.  Nie et al. \cite{AMGL} reformulated the standard spectral learning model and proposed a multi-view graph learning framework with adaptive weight learning. Liang et al. \cite{liangICDM,liangTNNLS}  simultaneously and explicitly modeled the consistency and inconsistency in multiple single-view graphs and fused them via an efficient optimization algorithm.

In terms of the computational complexity, some bipartite graph-based methods \cite{MVSC,LMVSC,MSGL,SMVSC,FPMVS-CAG,Huang2023} have been proposed, which, instead of building an $n\times n$ graph, resort to an $n\times m$ bipartite graph, where $m\ll n$ is the number of anchors, so as to make them computationally feasible for large-scale datasets. Nie et al. \cite{co-clustering-nie2017} aimed to learn an optimal bipartite graph with ideal connected components for co-clustering. Further, Nie et al. \cite{nie2021learning} utilized a dictionary matrix to learn a structured optimal bipartite graph via subspace clustering with the Laplacian rank constraint, which is designed for the single-view situation. For the multi-view situations, Sun et al. \cite{SMVSC} proposed a unified bipartite graph learning model with view-unified and orthogonal anchors, where view-specific weights are adaptively learned. In addition, the bipartite graph structure can also be utilized to reconstruct point-to-point relationships for the incomplete MVC problem \cite{guo2019anchors}.

In terms of the direct (or discrete) clustering, these aforementioned methods, such as LMVSC \cite{LMVSC} and FPMVS-CAG \cite{FPMVS-CAG}, mostly learn or build a unified graph and then conduct additional spectral partitioning on the graph to obtain the final clustering. However, the spectral partitioning (i.e., spectral clustering) often requires performing the $k$-means clustering on the spectral embeddings, which may bring in performance degradation due to the instability of $k$-means. To this end, some previous works incorporate the Laplacian rank constraint to learn a graph with discrete cluster structure, which can directly lead to the clustering result without performing additional partitioning \cite{MLAN,SwMC,MSGL,MVCTM,SFMC,nie2021learning,MCGC}. For example, Nie et al. \cite{SwMC} sought to learn a Laplacian rank constrained graph by considering the similarity matrices of multiple views. Kang et al. \cite{MSGL} performed the multi-view bipartite graph learning with the connectivity constraints to directly learn the clustering structure. But these MVC methods still lack the ability to unify the single-view and the unified bipartite graph learning in a joint learning model. To remedy these issues, in this paper, we aim to establish a unified framework with joint view-specific and view-consensus bipartite graph learning as well as discrete cluster structure learning for large-scale multi-view data.

\section{Methodology}\label{sec:method}
In this section, we present the formulation of the proposed UDBGL approach. Specifically, Section~\ref{sec:notations} introduces the main notations used throughout the paper. The view-specific subspace learning with anchors is represented in Section~\ref{sec:subspace_learning}. The Laplacian rank-constrained bipartite graph fusion is formulated in Section~\ref{sec:rank}. Finally, Section \ref{sec:UDBGL} describes the overall framework of UDBGL.

\subsection{Notations}\label{sec:notations}
Throughout this paper, we denote the scalar values by italic letters, the vectors by boldface lower-case letters, and the matrices by boldface capital letters. The $j$-th column of matrix $\textbf{M}$ is written as $\textbf{m}_{:j}$ (or $\textbf{M}(:,j)$), with its $i$-th entry being $m_{ij}$ (or $\textbf{M}(i,j)$). $\|\textbf{v}\|_2$ denotes the $2$-norm of a vector \textbf{v}, and $\|\textbf{M}\|_F$ denotes the Frobenius norm of matrix \textbf{M}. $\textbf{v}\ge 0$ (or $\textbf{M}\ge 0$) represents that all of the entries in this vector (or this matrix) are larger than or equal to zero. $\textbf{I}$ denotes the identity matrix. $\textbf{1}$ denotes a column vector with all entries being one. For clarity, Table~\ref{tab:notations} provides the frequently-used notations and their descriptions.

\begin{table}[!t]
\begin{center}
\centering
\caption{Notations} 
\label{tab:notations}
\begin{tabular}{m{2.5cm}m{5.5cm}}
\toprule
Notations &Descriptions\\
\midrule
$n$                                     &The number of samples\\
$c$                                     &The number of clusters\\
$V$                                     &The number of views\\
$m$                                     &The number of anchors\\
$d^{(v)}$                               &The number of features in the $v$-th view\\
$\textbf{X}^{(v)}\in\mathbb{R}^{d^{(v)}\times n}$       &Data matrix in the $v$-th view\\
$\textbf{Z}^{(v)}\in\mathbb{R}^{n\times m}$ &View-specific bipartite graph of the $v$-th view\\
$\textbf{P}\in\mathbb{R}^{n\times m}$       &View-consensus bipartite graph\\
$\textbf{A}^{(v)}\in\mathbb{R}^{d^{(v)}\times m}$             &Anchor matrix in the $v$-th view\\
$\bm\delta\in\mathbb{R}^{V}$                   &The view coefficient vector\\ &$\bm\delta=[\delta^{(1)},\delta^{(2)},\cdots,\delta^{(V)}]^\top$ \\
\bottomrule
\end{tabular}
\end{center}
\end{table}

\subsection{View-specific Subspace Learning with Anchors}\label{sec:subspace_learning}
Let a single-view data matrix be denoted as $\textbf{X}\in\mathbb{R}^{d\times n}$, which consists of $n$ samples with $d$ features. According to sparse subspace clustering (SSC) \cite{vidal2011subspace}, it is assumed that a data point can be written as a linear or affine combination of all other data points, i.e., $\textbf{X}=\textbf{X}\textbf{Z}^\top+\textbf{E}$, where $\textbf{Z}\in\mathbb{R}^{n\times n}$ is the self-representation matrix with the constraint $z_{ii}=0$ to enforce that each point cannot be represented by itself, and $\textbf{E}\in\mathbb{R}^{d\times n}$ is the error term. It aims to minimize the error term to find a better self-representation matrix, whose objective function can be written as
\begin{align}
\min_{\textbf{Z}}~&\left\|\textbf{X}-\textbf{X}\textbf{Z}^\top\right\|_F^2+\alpha\left\|\textbf{Z}\right\|_F^2,\notag\\
s.t.~&\textbf{Z}\ge0,\textbf{Z}\textbf{1}=\textbf{1};\forall i,z_{ii}=0,
\end{align}
where $\left\|\textbf{Z}\right\|_F^2$ is the regularization term that prevents the model from the trivial solution, and $\alpha>0$ is a hyper-parameter to control the influence of the regularization term. By constraining $\textbf{Z}\ge0,\textbf{Z}\textbf{1}=\textbf{1}$, the elements in \textbf{Z} with $0\le z_{ij}\le1$ can represent the similarity between the $i$-th sample and the $j$-th sample. Thus $\textbf{Z}$ can also be viewed as a similarity graph. Note that the elements in the $i$-th row of $\textbf{Z}$ denote the linear weighting coefficients of the $i$-th sample  reconstructed by the other sample points, i.e., $\textbf{x}_{:i}\approx z_{i1}\textbf{x}_{:1}+z_{i2}\textbf{x}_{:2}+\cdots+z_{in}\textbf{x}_{:n}$. Here, the row constraint (via $\textbf{Z}\textbf{1}=\textbf{1}$) is utilized to ensure that each row of $\textbf{Z}$ has at least one non-zero element.

For multi-view subspace clustering, it is challenging to fuse information from different views due to the potential inconsistency among multiple views \cite{gao2015multi}. Let $\textbf{X}^{(v)}\in\mathbb{R}^{d^{(v)}\times n}$ denote the $v$-th data matrix of the multi-view dataset, where $d^{(v)}$ is the number of features in the $v$-th view. Thus, we can perform the subspace learning on every single view as
\begin{align}
\label{eq:obj1-2}
\min_{\textbf{Z}^{(v)}}~&\sum_{v=1}^{V}\left\{\left\|\textbf{X}^{(v)}-\textbf{X}^{(v)}(\textbf{Z}^{(v)})^\top\right\|_F^2+\alpha\left\|\textbf{Z}^{(v)}\right\|_F^2\right\},\notag\\
&+\beta\mathcal{F}\left(\textbf{Z}^{(1)},\cdots,\textbf{Z}^{(V)}\right),\notag\\
s.t.~&\textbf{Z}^{(v)}\ge0,\textbf{Z}^{(v)}\textbf{1}=\textbf{1};\forall v,i,z_{ii}^{(v)}=0,
\end{align}
where $\mathcal{F}(\textbf{Z}^{(1)},\cdots,\textbf{Z}^{(V)})$ is a regularization term which enforces the consistency of multiple self-representation matrices $\textbf{Z}^{(v)}\in\mathbb{R}^{n\times n}$, and $\beta>0$ is a hyper-parameter to control the influence of the regularization term.

The above-mentioned model needs to learn multiple $n\times n$ graphs, which takes at least $\mathcal{O}(n^2)$ time. In the general subspace learning, it is assumed that a sample can be written as a linear or affine combination of all other samples (as shown in the objective~(\ref{eq:obj1-2})). For the self-expressiveness property, there may be much redundant information in the original large-scale dataset. Instead of using the entire dataset as the dictionary, using a smaller number of samples as the dictionary may significantly reduce the computational cost while maintaining sufficient representation ability for the latent subspaces. Here, the small number of representative samples are referred to as the anchors. Recently, it has proved to be an efficient and effective strategy to utilize the anchor graph (which is a bipartite graph between the orginal samples and the anchors) rather than the $n\times n$ full graph for large-scale multi-view datasets \cite{LMVSC,SFMC}. Specifically, a straightforward way to construct the anchors is to conduct the $k$-means clustering on the concatenated features and then split the obtained centers (i.e., anchors) of the concatenated features by views \cite{SDAGF}.
Let $[\textbf{A}^{(1)};\textbf{A}^{(2)};\cdots;\textbf{A}^{(V)}]\in\mathbb{R}^{\sum_vd^{(v)}\times m}$  denote the anchor matrix for the dataset $[\textbf{X}^{(1)};\textbf{X}^{(2)};\cdots;\textbf{X}^{(V)}]\in\mathbb{R}^{\sum_vd^{(v)}\times n}$, where $\textbf{A}^{(v)}$ is the anchor matrix for the $v$-th view and $m$ is the number of anchors. We extend the objective \eqref{eq:obj1-2} from the conventional subspace learning to the anchor-based subspace learning by replacing the data matrix $\textbf{X}^{(v)}$ with the anchor matrix $\textbf{A}^{(v)}$, and thus rewrite the objective function as
\begin{align}
\label{eq:obj1-3}
\min_{\textbf{Z}^{(v)}}~&\sum_{v=1}^{V}\left\{\left\|\textbf{X}^{(v)}-\textbf{A}^{(v)}(\textbf{Z}^{(v)})^\top\right\|_F^2+\alpha\left\|\textbf{Z}^{(v)}\right\|_F^2\right\},\notag\\
&+\beta\mathcal{F}\left(\textbf{Z}^{(1)},\cdots,\textbf{Z}^{(V)}\right),\notag\\
&s.t.~\textbf{Z}^{(v)}\ge0,\textbf{Z}^{(v)}\textbf{1}=\textbf{1},
\end{align}
where the representation matrix $\textbf{Z}^{(v)}\in\mathbb{R}^{n\times m}$ can be interpreted as a bipartite graph between the $n$ original samples and the $m$ anchors.

\subsection{Rank-constrained Bipartite Graph Fusion}\label{sec:rank}
In this section, we proceed to fuse multiple bipartite graphs from multiple views into a unified bipartite graph. Notably, the multiple bipartite graphs are associated with adaptively-learned weights, and a Laplacian rank constraint is further imposed for learning a unified bipartite graph with discrete cluster structure.

Formally, the objective function for bipartite graph fusion can be formulated as
\begin{align}
&\min_{\textbf{P},\bm\delta}~\left\|\sum_{v=1}^V\delta^{(v)}\textbf{Z}^{(v)}-\textbf{P}\right\|_F^2,\notag\\
&s.t.~\bm\delta^\top\textbf{1}=1,\bm\delta\ge0;\textbf{P}\in\bm\Omega,
\end{align}
where $\bm\delta=[\delta^{(1)},\delta^{(2)},\cdots,\delta^{(V)}]^\top$ is the view coefficient vector, and $\bm\Omega=\left\{\textbf{P}\mid\textbf{P}\textbf{1}=\textbf{1},\textbf{P}\ge0\right\}$ ensures that $\textbf{Z}^{(v)}$ and $\textbf{P}$ are at a similar scale.

After obtaining the unified bipartite graph $\textbf{P}$, the traditional strategy is to approximately construct a similarity graph among all samples through the obtained bipartite graph.
The doubly-stochastic similarity graph $\textbf{S}\in\mathbb{R}^{n\times n}$ \cite{liu2010large,guo2019anchors}, whose summation of each row or each column equals to $1$, can be built as
\begin{align}
\label{eq:post-processing}
\textbf{S}=&\textbf{P}\bm\Lambda^{-1}\textbf{P}^\top,\\
\bm\Lambda=&diag\left(\textbf{P}^\top\textbf{1}\right)\in\mathbb{R}^{m\times m},
\end{align}
where the $diag\left(\cdot\right)$ function returns a diagonal matrix whose main diagonal elements are the corresponding elements of input vector. Subsequently, the spectral clustering can be conducted on the similarity graph $\textbf{S}$, which takes at least $\mathcal{O}(n^2c)$ time \cite{LMVSC}, where $c$ is the desired number of clusters. Regarding this, we redefine the similarity graph $\textbf{S}$ in a different way from the aforementioned Eq. \eqref{eq:post-processing}. That is
 \begin{align}
\textbf{S}=\begin{pmatrix}
&\textbf{P}\\
\textbf{P}^\top&\\
\end{pmatrix}\in\mathbb{R}^{(n+m)\times(n+m)}.
\end{align}

\begin{lemma}
\label{le:rank-constraint}
\cite{von2007tutorial,SFMC} Let $\textbf{S}$ be an undirected graph with non-negative elements. The multiplicity of the eigenvalue $0$ of the normalized Laplacian matrix $\widetilde{\textbf{L}}_\textbf{S}$ equals the number of connected components in the graph $\textbf{S}$.
\end{lemma}

In order to reduce the time complexity and directly achieve the discrete cluster structure without additional partitioning, we impose a rank constraint on the normalized Laplacian matrix (i.e., $rank(\widetilde{\textbf{L}}_\textbf{S})=n+m-c$) of the similarity graph $\textbf{S}$, where $\widetilde{\textbf{L}}_\textbf{S}=\textbf{I}-\textbf{D}_\textbf{S}^{-\frac{1}{2}}\textbf{S}\textbf{D}_\textbf{S}^{-\frac{1}{2}}$ is the normalized Laplacian matrix, and $\textbf{D}_\textbf{S}=diag(\textbf{S}\textbf{1})\in\mathbb{R}^{(n+m)\times(n+m)}$ is the degree matrix. According to Lemma \ref{le:rank-constraint}, the rank constraint enforces that the multiplicity of eigenvalue $0$ of $\widetilde{\textbf{L}}_\textbf{S}$ should be $c$, which leads to the learned graph $\textbf{S}$ with exact $c$ connected components.

Further, since $\textbf{S}$ and $\textbf{P}$ represent the same graph, the $c$-connected components of the bipartite graph $\textbf{P}$ are guaranteed through the $c$-connected graph $\textbf{S}$. Thus the constraints on $\textbf{P}$ can be rewritten as
\begin{align}\bm\Omega=\{\textbf{P}\mid\textbf{P}\textbf{1}=\textbf{1},\textbf{P}\ge0;rank(\widetilde{\textbf{L}}_\textbf{S})=n+m-c\}.
\end{align}

Through the Laplacian rank constraint, the final clustering result can be directly gained from the bipartite graph $\textbf{P}$ without additional partitioning. In particular, the bipartite graph $\textbf{P}$ with the Laplacian low rank constraint will have a specific number of connected components \cite{von2007tutorial,SFMC}. By treating the samples in the same connected component as a cluster, the clustering result can then be obtained. Thereby, we rewrite the objective function of the bipartite graph fusion as follows:
\begin{align}
\label{eq:bipartite_graph_fusion}
\min_{\textbf{P},\bm\delta}~&\left\|\sum_{v=1}^V\delta^{(v)}\textbf{Z}^{(v)}-\textbf{P}\right\|_F^2.\notag\\
s.t.~\bm\delta^\top\textbf{1}=1,\bm\delta\ge0;&\textbf{P}\textbf{1}=\textbf{1},\textbf{P}\ge0;rank(\widetilde{\textbf{L}}_\textbf{S})=n+m-c.
\end{align}

Both the general graph and the bipartite graph can utilize the rank constraint to learn the discrete cluster structure, but they are quite different in the connectivity and optimization. In terms of the connectivity, the general graph constructs different connected components through the similarity between data samples, while in a bipartite graph it should be decided through the anchors whether two data samples in the bipartite graph are connected. That is, the connectivity between two samples should be achieved through one or more anchors. In terms of the optimization, the discrete cluster structure learning of a general graph is usually more time-consuming, while utilizing the bipartite graph structure can significantly reduce the computational complexity.

\subsection{Multi-view Clustering via Unified and Discrete Bipartite Graph Learning}
\label{sec:UDBGL}
Though previous MVC methods based on bipartite graph learning have achieved considerable progress, yet most of them \textit{either} pre-construct the view-specific bipartite graphs and then learn a unified bipartite graph \cite{co-clustering-nie2017,SDAGF,SFMC,MVSC} \textit{or} learn multiple view-specific bipartite graphs and then simply concatenate the learned view-specific bipartite graphs into a unified one \cite{LMVSC}. However, the pre-constructed bipartite graphs lack the ability of adaptive learning and refining, while the simple concatenation cannot well explore the latent structures of multiple bipartite graphs. Surprisingly, there remains a challenging gap between the view-specific bipartite graph learning and the view-consensus bipartite graph learning in most of previous works \cite{co-clustering-nie2017,SDAGF,SFMC,MVSC,LMVSC}.

In this paper, we seek to unify the view-specific bipartite graph learning, the view-consensus bipartite graph learning,  and the discrete cluster structure learning in a joint learning model. Particularly, the objective~(\ref{eq:obj1-3}) (for anchor-based subspace learning) and the objective~(\ref{eq:bipartite_graph_fusion}) are jointly formulated, where the graph fusion objective serves as a regularization term for the objective~\eqref{eq:obj1-3} and the Laplacian rank constraint bridges the gap between the bipartite graph learning and the discrete cluster structure learning.
Formally, the objective function for the unified and discrete bipartite graph learning can be formulated as follows:
\begin{align}
\label{eq:obj-final}
\min_{\textbf{Z}^{(v)},\textbf{P},\bm\delta}~&\sum_{v=1}^{V}\left\{\left\|\textbf{X}^{(v)}-\textbf{A}^{(v)}(\textbf{Z}^{(v)})^\top\right\|_F^2+\alpha\left\|\textbf{Z}^{(v)}\right\|_F^2\right\}\notag\\
&+\beta\left\|\sum_{v=1}^V\delta^{(v)}\textbf{Z}^{(v)}-\textbf{P}\right\|_F^2,\notag\\
s.t.~&\textbf{Z}^{(v)}\ge0,\textbf{Z}^{(v)}\textbf{1}=\textbf{1};\bm\delta^\top\textbf{1}=1,\bm\delta\ge0;\textbf{P}\textbf{1}=\textbf{1},\textbf{P}\ge0;\notag\\
&rank(\widetilde{\textbf{L}}_\textbf{S})=n+m-c.
\end{align}

In the next section, we will present a fast optimization algorithm to minimize this unified objective function, which enables the mutual promotion of the view-specific bipartite graphs and the view-consensus bipartite graph, and direct achieves the final clustering result without requiring additional graph partitioning.

\section{Optimization and Theoretical Analysis}\label{sec:opt}
In this section, we design an efficient alternating optimization algorithm to minimize the objective function~(\ref{eq:obj-final}). Specifically, in each iteration, we update each of the variables $\textbf{P}$, $\textbf{Z}^{(v)}$, and $\bm\delta$ while fixing the other variables in Sections \ref{sec:UpP}, \ref{sec:UpZ} and \ref{sec:UpDelta}, respectively. Further, we provide the computational complexity analysis and the theoretical convergence analysis in Sections \ref{sec:time} and \ref{sec:convergence}, respectively.

\subsection{Update $\textbf{P}$}
\label{sec:UpP}
With the other variables fixed, the subproblem that only relates to $\textbf{P}$ can be written as
\begin{align}
\label{eq:Up_P1}
&\min_{\textbf{P}}~\left\|\sum_{v=1}^V\delta^{(v)}\textbf{Z}^{(v)}-\textbf{P}\right\|_F^2,\notag\\
s.t.~~\textbf{P}\textbf{1}&=\textbf{1},\textbf{P}\ge0;rank(\widetilde{\textbf{L}}_\textbf{S})=n+m-c.
\end{align}
According to \cite{SFMC} and the Ky Fan's Theorems \cite{fan1949theorem}, let $\textbf{B}=\sum_{v=1}^V\delta^{(v)}\textbf{Z}^{(v)}$, the subproblem \eqref{eq:Up_P1} can be further transformed into
\begin{align}
\label{eq:Up_P2}
\min_{\textbf{P},\textbf{F}}~&\left\|\textbf{B}-\textbf{P}\right\|_F^2+\gamma ~Tr\left(\textbf{F}^\top\widetilde{\textbf{L}}_\textbf{S}\textbf{F}\right),\notag\\
s.t.~~&\textbf{P}\textbf{1}=\textbf{1},\textbf{P}\ge0;\textbf{F}^\top\textbf{F}=\textbf{I},
\end{align}
where $\textbf{F}\in\mathbb{R}^{(n+m)\times c}$ is the indicator matrix, and $\gamma$ can be automatically determined based on the number of connected components of $\textbf{P}$. Further, we  alternately optimize the variables $\textbf{F}$ and $\textbf{P}$.

First, with $\textbf{P}$ fixed, the subproblem w.r.t. $\textbf{F}$ can be formulated as
\begin{align}
\label{eq:Up_F1}
\min_{\textbf{F}^\top\textbf{F}=\textbf{I}}~&Tr(\textbf{F}^\top\widetilde{\textbf{L}}_\textbf{S}\textbf{F})\notag\\
\Leftrightarrow\max_{\textbf{F}_{(n)}^\top\textbf{F}_{(n)}+\textbf{F}_{(m)}^\top\textbf{F}_{(m)}=\textbf{I}}&Tr(\textbf{F}_{(n)}^\top\textbf{D}_{(n)}^{-\frac{1}{2}}\textbf{P}\textbf{D}_{(m)}^{-\frac{1}{2}}\textbf{F}_{(m)}),
\end{align}
with
\begin{align}
\textbf{F}&=\begin{pmatrix}
\textbf{F}_{(n)}\\
\textbf{F}_{(m)}\\
\end{pmatrix},~\textbf{F}_{(n)}\in\mathbb{R}^{n\times c},\textbf{F}_{(m)}\in\mathbb{R}^{m\times c},\\
\textbf{D}_\textbf{S}&=\begin{pmatrix}
\textbf{D}_{(n)}&\\
&\textbf{D}_{(m)}\\
\end{pmatrix},~\textbf{D}_{(n)}\in\mathbb{R}^{n\times n},\textbf{D}_{(m)}\in\mathbb{R}^{m\times m},
\end{align}
According to \cite{co-clustering-nie2017}, the optimal solutions to the problem \eqref{eq:Up_F1} can be obtained as $\textbf{F}_{(n)}=\frac{\sqrt{2}}{2}\textbf{U}$ and $\textbf{F}_{(m)}=\frac{\sqrt{2}}{2}\textbf{V}$, where $\textbf{U}$ and $\textbf{V}$ are the largest $c$ left and right singular vectors of $\textbf{D}_{(n)}^{-\frac{1}{2}}\textbf{P}\textbf{D}_{(m)}^{-\frac{1}{2}}$, respectively.

Second, according to \cite{SFMC}, we can have the following equality:
\begin{align}
Tr(\textbf{F}^\top\widetilde{\textbf{L}}_\textbf{S}\textbf{F})=\sum_{i=1}^n\sum_{j=1}^m\left\|\frac{\textbf{F}_{(n)}(i,:)}{\sqrt{\textbf{D}_{(n)}(i,i)}}-\frac{\textbf{F}_{(m)}(j,:)}{\sqrt{\textbf{D}_{(m)}(j,j)}}\right\|_2^2p_{ij}.
\end{align}
Let $q_{ij}=\left\|\frac{\textbf{F}_{(n)}(i,:)}{\sqrt{\textbf{D}_{(n)}(i,i)}}-\frac{\textbf{F}_{(m)}(j,:)}{\sqrt{\textbf{D}_{(m)}(j,j)}}\right\|_2^2$. With $\textbf{F}$ fixed, the subproblem w.r.t. $\textbf{P}$ can be written as
\begin{align}
\label{eq:Up_P3}
\min_{\textbf{P}\textbf{1}=\textbf{1},\textbf{P}\ge0}\sum_{i=1}^n\sum_{j=1}^m\left(b_{ij}-p_{ij}\right)^2+\gamma q_{ij}p_{ij}.
\end{align}
Because the optimization of each row of $\textbf{P}$ in the subproblem \eqref{eq:Up_P3} is independent, we optimize $\textbf{P}$ by rows:
\begin{align}
\label{eq:Up_P4}
\min_{\textbf{p}_{i:}\textbf{1}=1,\textbf{p}_{i:}\ge0}\left\|\textbf{p}_{i:}-(\textbf{b}_{i:}-\frac{\gamma}{2}\textbf{q}_{i:})\right\|_2^2.
\end{align}
Then the closed form solution $\textbf{p}_{i:}$ can be obtained according to \cite{Anew}.

\subsection{Update $\textbf{Z}^{(v)}$}
\label{sec:UpZ}
With the other variables fixed, the subproblem that only relates to $\textbf{Z}^{(v)}$ can be written as
\begin{align}
\label{eq:Up_Z1}
\min_{\textbf{Z}^{(v)}}~&\left\|\textbf{X}^{(v)}-\textbf{A}^{(v)}(\textbf{Z}^{(v)})^\top\right\|_F^2+\alpha\left\|\textbf{Z}^{(v)}\right\|_F^2\notag\\
&+\beta\left\|\sum_{i=1}^V\delta^{(i)}\textbf{Z}^{(i)}-\textbf{P}\right\|_F^2,\notag\\
s.t.~&~\textbf{Z}^{(v)}\ge0,\textbf{Z}^{(v)}\textbf{1}=\textbf{1}.
\end{align}
Since the optimization of each row of $\textbf{Z}^{(v)}$ in the subproblem \eqref{eq:Up_Z1} is independent, we optimize $\textbf{Z}^{(v)}$ by rows:
\begin{align}
\label{eq:Up_Z2}
\min_{\textbf{z}^{(v)}_{j:}}~&\left\|\textbf{x}^{(v)}_{:j}-\textbf{A}^{(v)}(\textbf{z}^{(v)}_{j:})^\top\right\|_2^2+\alpha\left\|\textbf{z}^{(v)}_{j:}\right\|_2^2\notag\\
&+\beta\left\|\sum_{i=1}^V\delta^{(i)}\textbf{z}^{(i)}_{j:}-\textbf{p}_{j:}\right\|_2^2,\notag\\
s.t.~&~\textbf{z}^{(v)}_{j:}\ge0,\textbf{z}^{(v)}_{j:}\textbf{1}=1.
\end{align}
The above optimization problem \eqref{eq:Up_Z2} can be easily formulated as the following quadratic programming (QP) problem
\begin{align}
\label{eq:Up_Z3}
\min_{\textbf{z}^{(v)}_{j:}}~&\textbf{z}^{(v)}_{j:}{\bar{\textbf{H}}}(\textbf{z}^{(v)}_{j:})^\top+\textbf{z}^{(v)}_{j:}{\bar{\textbf{f}}},\notag\\
s.t.~&~\textbf{z}^{(v)}_{j:}\ge0,\textbf{z}^{(v)}_{j:}\textbf{1}=1,
\end{align}
where
\begin{align}
{\bar{\textbf{H}}}=&(\textbf{A}^{(v)})^\top\textbf{A}^{(v)}+\left(\alpha+\beta(\delta^{(v)})^2\right)\textbf{I},\notag\\
{\bar{\textbf{f}}}=-2(\textbf{A}^{(v)})^\top&\textbf{x}^{(v)}_{:j}+2\beta\delta^{(v)}\left(\sum_{i\neq v}\delta^{(i)}(\textbf{z}^{(i)}_{j:})^\top-\textbf{p}_{j:}^\top\right).
\end{align}
The problem~\eqref{eq:Up_Z3} can be solved by using the augmented Lagrangian multiplier (ALM) \cite{ALM} (please see Section \ref{sec:UpDelta} for the details).

\begin{algorithm}
	\renewcommand{\algorithmicrequire}{\textbf{Input:}}
	\renewcommand{\algorithmicensure}{\textbf{Preparation:}}
	\caption{Multi-view clustering via unified and discrete bipartite graph learning (UDBGL)}\label{algorithm}
	\begin{algorithmic}[1]
		\REQUIRE Multi-view dataset $\{\textbf{X}^{(v)}\}_{v=1}^V$, the number of clusters $c$, the number of anchors $m\ge c$, the parameters $\alpha>0$ and $\beta>0$.
		\ENSURE Normalize $\{\textbf{X}^{(v)}\}_{v=1}^V$. Conduct $k$-means on the concatenated features $[\textbf{X}^{(1)};\textbf{X}^{(2)};\cdots;\textbf{X}^{(V)}]$ to obtain the anchors $[\textbf{A}^{(1)};\textbf{A}^{(2)};\cdots;\textbf{A}^{(V)}]$.
		\renewcommand{\algorithmicensure}{\textbf{Initialization:}}
		\ENSURE  $\bm\delta=\textbf{1}/V$. Initialize $\textbf{Z}^{(v)}$ by $K$-nearest-neighbor ($K$-NN) bipartite graph.
		\REPEAT
		\STATE Initialize $\gamma=0.1$, $\textbf{D}_{(n)}=diag(\textbf{B}\textbf{1})$, and $\textbf{D}_{(m)}=diag(\textbf{B}^\top\textbf{1})$.
		\STATE Initialize $\textbf{F}$ by solving problem \eqref{eq:Up_F1} (via replacing $\textbf{P}$ with $\textbf{B}$).
		\REPEAT
		\STATE Update $q_{ij}=\left\|\frac{\textbf{F}_{(n)}(i,:)}{\sqrt{\textbf{D}_{(n)}(i,i)}}-\frac{\textbf{F}_{(m)}(j,:)}{\sqrt{\textbf{D}_{(m)}(j,j)}}\right\|_2^2$.
		\STATE Update $\textbf{P}$ by solving problem \eqref{eq:Up_P4} and update $\textbf{D}_\textbf{S}$.
		\STATE Update $\textbf{F}$ by solving problem \eqref{eq:Up_F1}.
		\STATE $temp =~$the multiplicity of the eigenvalue $0$ of $\widetilde{\textbf{L}}_\textbf{S}$.
		\IF{$temp<c$}
		\STATE $\gamma\leftarrow2\gamma$.
		\ELSIF{$temp>c$}
		\STATE $\gamma\leftarrow\gamma/2$.
		\ENDIF
		\UNTIL{Obtaining the $c$-connected bipartite graph $\textbf{P}$}
		\STATE $\forall v$, update $\textbf{Z}^{(v)}$ by solving problem \eqref{eq:Up_Z1}.
		\STATE Calculate $\textbf{H}=\hat{\textbf{Z}}^\top\hat{\textbf{Z}}$, and $\textbf{f}=2\hat{\textbf{Z}}^\top\hat{\textbf{p}}$.
		\STATE Initialize $\bm\delta=\textbf{1}/V$, $\mu=2$, and $\bm\eta$.
		\REPEAT
		\STATE Update $\bm\rho$ by Eq. \eqref{eq:Up_delta6}.
		\STATE Update $\bm\delta$ by solving problem \eqref{eq:Up_delta5}.
		\STATE $\bm\eta\leftarrow\bm\eta+\mu(\bm\delta-\bm\rho)$,$\mu\leftarrow2\mu$.
		\UNTIL{$\bm\delta$ converges}
		\UNTIL{Convergence or maximum iteration reached}
		\renewcommand{\algorithmicensure}{\textbf{Output:}}
		\ENSURE  The final clustering can be directly obtained from $\textbf{P}$.
		
	\end{algorithmic}
\end{algorithm}

\begin{table*}[!th]
	\centering
	\caption{Details of the multi-view datasets in our experiments}
	\label{tab:datasets}
	\begin{tabular}{m{2cm}<{\centering}|m{1cm}<{\centering}m{1cm}<{\centering}m{1cm}<{\centering}m{7cm}<{\centering}}
		\toprule
		Dataset &\#Sample    &\#View  &\#Class  &Dimension\\
		\midrule
		WebKB-Texas     &187    &2 &5   &Citation(187), Content(1,703)\\
		MSRCv1          &210    &4 &7   &CM(24), GIST(512), LBP(256), GENT(254)\\
		Out-Scene       &2,688  &4 &8   &GIST(512), HOG(432), LBP(256), Gabor(48)\\
		Cora            &2,708  &4 &7   &View1(2,708), View2(1,433), View3(2,708), View4(2,708)\\
		Citeseer        &3,312  &2 &6   &Content(3,703), Citation(3,312)\\
        {Notting-Hill}        &4,660  &3  &5  &Gabor(6,750),LBP(3,304),Intensity(2,000)\\
		VGGFace        &34,027 &4 &50  &LBP(944), HOG(576), GIST(512), Gabor(640)\\
        {YoutubeFaces}          &51,615 &4 &15  &LBP(944), HOG(576), GIST(512), Gabor(640)\\
		CIFAR-10        &60,000 &4 &10  &LBP(944), HOG(576), GIST(512), Gabor(640)\\
		CIFAR-100       &60,000 &4 &100 &LBP(944), HOG(576), GIST(512), Gabor(640)\\
		\bottomrule
	\end{tabular}
\end{table*}

\subsection{Update $\bm\delta$}
\label{sec:UpDelta}
With the other variables fixed, the subproblem that only relates to $\bm\delta$ can be written as
\begin{align}
\label{eq:Up_delta1}
\min_{\bm\delta}~&\left\|\sum_{v=1}^V\delta^{(v)}\textbf{Z}^{(v)}-\textbf{P}\right\|_F^2,\notag\\
s.t.~~&\bm\delta^\top\textbf{1}=1,\bm\delta\ge0.
\end{align}
In order to solve the subproblem \eqref{eq:Up_delta1}, we vectorize each matrix $\textbf{Z}^{(v)}$ into a vector $\hat{\textbf{z}}^{(v)}$, that is
\begin{align}
\label{eq:vectorization}
\hat{\textbf{z}}^{(v)}=\left[\textbf{z}^{(v)}_{:1};\textbf{z}^{(v)}_{:2};\cdots;\textbf{z}^{(v)}_{:m}\right]\in\mathbb{R}^{nm\times1},
\end{align}
where $\textbf{z}^{(v)}_{:i}$ represents the $i$-th column of $\textbf{Z}^{(v)}$. Further, we stack the vectors $\hat{\textbf{z}}^{(v)}$ for $V$ views into a matrix $\hat{\textbf{Z}}=[\hat{\textbf{z}}^{(1)},\hat{\textbf{z}}^{(2)},\cdots,\hat{\textbf{z}}^{(V)}]\in\mathbb{R}^{nm\times V}$. Note that $\textbf{B}=\sum_{v=1}^V\delta^{(v)}\textbf{Z}^{(v)}$, and $\hat{\textbf{b}}$ is the vectorization of $\textbf{B}$ (similar to Eq. \eqref{eq:vectorization}), so we have $\hat{\textbf{b}}=\hat{\textbf{Z}}\bm\delta$. Similarly, $\hat{\textbf{p}}$ is the vectorization of $\textbf{P}$, then the problem \eqref{eq:Up_delta1} can be rewritten as
\begin{align}
\label{eq:Up_delta2}
&\min_{\bm\delta^\top\textbf{1}=1,\bm\delta\ge0}\left\|\hat{\textbf{Z}}\bm\delta-\hat{\textbf{p}}\right\|_2^2\nonumber\\ \Leftrightarrow &\min_{\bm\delta^\top\textbf{1}=1,\bm\delta\ge0}\bm\delta^\top\textbf{H}\bm\delta-\bm\delta^\top\textbf{f},
\end{align}
where $\textbf{H}=\hat{\textbf{Z}}^\top\hat{\textbf{Z}}$, and $\textbf{f}=2\hat{\textbf{Z}}^\top\hat{\textbf{p}}$. The objective \eqref{eq:Up_delta2} is a convex quadratic programming with the semi-definite quadratic matrix $\textbf{H}$. Here, we use the augmented Lagrangian multiplier (ALM) \cite{ALM} to optimize the objective \eqref{eq:Up_delta2}, whose solution is obtained by solving its equivalence:
\begin{align}
\label{eq:Up_delta3}
\min_{\bm\delta^\top\textbf{1}=1,\bm\delta\ge0,\bm\delta=\bm\rho}\bm\delta^\top\textbf{H}\bm\rho-\bm\delta^\top\textbf{f}.
\end{align}
The augmented Lagrangian function of the objective \eqref{eq:Up_delta3} can be written as:
\begin{align}
\label{eq:Up_delta4}
\min_{\bm\delta^\top\textbf{1}=1,\bm\delta\ge0,\bm\rho}\bm\delta^\top\textbf{H}\bm\rho-\bm\delta^\top\textbf{f}+\frac{\mu}{2}\left\|\bm\delta-\bm\rho+\frac{1}{\mu}\bm\eta\right\|_2^2,
\end{align}
where the optimal solutions $\bm\delta$ and $\bm\rho$ can be optimized alternately. The third term of the objective \eqref{eq:Up_delta4} is a penalty term which ensures $\bm\delta=\bm\rho$. The penalty factor $\mu>0$ is gradually increased in each iteration. Meanwhile, $\bm\eta\in\mathbb{R}^{V}$ can be updated by $\bm\eta\leftarrow\bm\eta+\mu(\bm\delta-\bm\rho)$.

First, to update $\bm\rho$ with $\bm\delta$ fixed, the Lagrange function w.r.t. $\bm\rho$ can be written as
\begin{align}
\mathcal{L}(\bm\rho)={\bm\delta^*}^\top\textbf{H}\bm\rho+\frac{\mu}{2}\left\|\bm\delta^*-\bm\rho+\frac{1}{\mu}\bm\eta\right\|_2^2,
\end{align}
where $\bm\delta$ is fixed with $\bm\delta^*$. By setting the derivative of $\mathcal{L}(\bm\rho)$ w.r.t. $\bm\rho$ to zero, we can obtain the optimal solution as
\begin{align}
\label{eq:Up_delta6}
\bm\rho^*=\bm\delta^*+\frac{1}{\mu}(\bm\eta-\textbf{H}^\top\bm\delta^*).
\end{align}

Second, to update $\bm\delta$ with fixed $\bm\rho^*$, the objective \eqref{eq:Up_delta4} can be transformed into solving the following problem
\begin{align}
\label{eq:Up_delta5}
\min_{\bm\delta^\top\textbf{1}=1,\bm\delta\ge0}\left\|\bm\delta-\bm\rho^*+\frac{1}{\mu}(\bm\eta+\textbf{H}\bm\rho^*-\textbf{f})\right\|_2^2.
\end{align}
Thus, the problem \eqref{eq:Up_delta5} can be solved with a closed form solution according to \cite{Anew}.

For clarity, the overall algorithm of UDBGL is described in Algorithm~ \ref{algorithm}.

\subsection{Computational Complexity Analysis}
\label{sec:time}
In this section, we analyze the computational complexity of the proposed algorithm. The computational cost of UDBGL involves the computation of the preparation, the initialization, and the optimization of three variables. First, the normalization and the $k$-means take $\mathcal{O}(nd)$ and $\mathcal{O}(ndmt_1)$ time, respectively, where $d=\sum_vd^{(v)}$ and $t_1$ is the number of $k$-means iterations. Second, the cost of updating $\textbf{P}$ is $\mathcal{O}(nmct_2+nm^2t_2+m^3t_2)$, where $t_2$ is the iteration number of updating $\textbf{P}$. Third, the cost of updating $\textbf{Z}^{(v)}$ (for all $v$) is $\mathcal{O}(nm^2d)$. Finally, the cost of updating $\bm\delta$ is $\mathcal{O}(V^2nm)$. Since $n\gg m$, and $c,V,t_2$ are small constants, the time complexity of UDBGL is linear in $n$.

\subsection{Convergence analysis}
\label{sec:convergence}
In this section, we analyze the convergence property of the proposed optimization algorithm. In Section \ref{sec:UpP}, we can obtain the optimal solutions for the subproblems of $\textbf{F}$ and $\textbf{P}$. In order to obtain a bipartite graph with a specific number of connected components, we dynamically adjust the parameter $\gamma$, which leads to a bipartite graph with discrete cluster structure but may undermine the monotonicity of the objective function of the subproblem~\eqref{eq:Up_P1}. For updating $\textbf{Z}^{(v)}$, the QP problem \eqref{eq:Up_Z3} can be proved to be convex. For any non-zero vector $\bar{\textbf{x}}$, we have
\begin{align}
\bar{\textbf{x}}^\top\bar{\textbf{H}}\bar{\textbf{x}}&=\bar{\textbf{x}}^\top\{(\textbf{A}^{(v)})^\top\textbf{A}^{(v)}+(\alpha+\beta(\delta^{(v)})^2)\textbf{I}\}\bar{\textbf{x}}\notag\\
&=||\textbf{A}^{(v)}\bar{\textbf{x}}||_2^2+(\alpha+\beta(\delta^{(v)})^2)||\bar{\textbf{x}}||_2^2\ge0.
\end{align}
Thus, the matrix of quadratic form is positive semi-definite, and the objective function of the QP problem \eqref{eq:Up_Z3} is convex. Note that the constraint functions are affine. Therefore, the QP problem \eqref{eq:Up_Z3} is convex. Similarly, updating $\bm\delta$ is also a convex optimization problem. To summarize, except for the dynamic adjustment of the parameter $\gamma$ in the subproblem~\eqref{eq:Up_P1}, the optimal solutions for all the other subproblems can achieved. In the following, the empirical convergence of UDBGL will also be verified in Section~\ref{sec:empirical_convergence}.

\section{Experiments}
\label{sec:experiments}
In this section, we evaluate the proposed UDBGL method against the state-of-the-art MVC methods on ten multi-view datasets. All the experiments are conducted on a PC with 16GB RAM and an Intel i5-6600 CPU.

\begin{table*}[!th]
	\centering
	\newcommand{\tabincell}[2]{\begin{tabular}{@{}#1@{}}#2\end{tabular}}
	\caption{Average NMI(\%) over 20 runs by different multi-view clustering methods. The best result is highlighted in \textbf{bold}.}
	\label{tab:NMI}
		\resizebox{\textwidth}{!}{ \begin{tabular}{m{1.7cm}<{\centering}m{1.2cm}<{\centering}m{1.2cm}<{\centering}m{1.2cm}<{\centering}m{1.2cm}<{\centering}m{1.2cm}<{\centering}m{1.2cm}<{\centering}m{1.2cm}<{\centering}m{1.2cm}<{\centering}m{1.2cm}<{\centering}m{1.2cm}<{\centering}m{1.2cm}<{\centering}m{1.2cm}<{\centering}}
\toprule
Datasets	&MVSC	&AMGL	&MLAN	&SwMC	&MVCTM	&SFMC	&LMVSC	&MSGL	&SMVSC	&FPMVS-CAG	&EOMSC-CA	&\textbf{UDBGL}	\\
\midrule
WebKB-Texas	&1.98$_{\pm0.25}$	&9.89$_{\pm2.61}$	&7.11$_{\pm1.72}$	&9.37$_{\pm0.00}$	&28.74$_{\pm0.00}$	&6.25$_{\pm0.00}$	&20.86$_{\pm0.00}$	&4.56$_{\pm4.49}$	&21.42$_{\pm0.00}$	&22.76$_{\pm0.00}$	&18.84$_{\pm0.00}$	&\textbf{37.14}$_{\pm0.00}$	\\
MSRCv1	&49.34$_{\pm3.78}$	&58.84$_{\pm6.33}$	&73.46$_{\pm0.37}$	&62.69$_{\pm0.00}$	&68.91$_{\pm0.00}$	&52.35$_{\pm0.00}$	&24.60$_{\pm0.00}$	&26.98$_{\pm9.18}$	&61.09$_{\pm0.00}$	&56.62$_{\pm0.00}$	&44.65$_{\pm0.00}$	&\textbf{77.07}$_{\pm0.00}$	\\
Out-Scene	&23.39$_{\pm14.73}$	&44.19$_{\pm3.96}$	&37.38$_{\pm0.00}$	&43.69$_{\pm0.00}$	&47.29$_{\pm0.00}$	&39.72$_{\pm0.00}$	&45.97$_{\pm0.00}$	&15.10$_{\pm8.52}$	&51.67$_{\pm0.00}$	&\textbf{53.04}$_{\pm0.00}$	&38.09$_{\pm0.00}$	&52.45$_{\pm0.00}$	\\
Cora	&0.25$_{\pm0.09}$	&9.00$_{\pm4.79}$	&2.55$_{\pm0.00}$	&8.69$_{\pm0.01}$	&5.47$_{\pm0.00}$	&4.79$_{\pm0.00}$	&12.82$_{\pm0.00}$	&0.54$_{\pm0.81}$	&28.73$_{\pm0.00}$	&18.19$_{\pm0.00}$	&6.56$_{\pm0.00}$	&\textbf{30.67}$_{\pm0.00}$	\\
Citeseer	&0.40$_{\pm0.11}$	&0.53$_{\pm0.06}$	&1.85$_{\pm4.51}$	&1.22$_{\pm0.00}$	&1.68$_{\pm0.00}$	&2.43$_{\pm0.00}$	&8.75$_{\pm0.00}$	&0.77$_{\pm1.20}$	&17.86$_{\pm0.00}$	&14.73$_{\pm0.00}$	&14.88$_{\pm0.00}$	&\textbf{24.06}$_{\pm0.00}$	\\
{Notting-Hill}	&53.98$_{\pm19.66}$	&8.19$_{\pm2.82}$	&55.50$_{\pm0.00}$	&8.77$_{\pm0.00}$	&47.95$_{\pm0.00}$	&84.11$_{\pm0.00}$	&68.06$_{\pm0.00}$	&65.20$_{\pm19.10}$	&66.64$_{\pm0.00}$	&67.15$_{\pm0.00}$	&75.13$_{\pm0.00}$	&\textbf{87.17}$_{\pm0.00}$	\\
VGGFace	&N/A	&N/A	&N/A	&N/A	&N/A	&7.79$_{\pm0.00}$	&13.05$_{\pm0.00}$	&2.49$_{\pm2.30}$	&11.16$_{\pm0.00}$	&12.33$_{\pm0.00}$	&12.20$_{\pm0.00}$	&\textbf{14.27}$_{\pm0.00}$	\\
{YoutubeFaces}	&N/A	&N/A	&N/A	&N/A	&N/A	&78.06$_{\pm0.00}$	&74.91$_{\pm0.00}$	&67.08$_{\pm2.83}$	&71.14$_{\pm0.00}$	&73.54$_{\pm0.00}$	&63.94$_{\pm0.00}$	&\textbf{78.69}$_{\pm0.00}$	\\
CIFAR-10	&N/A	&N/A	&N/A	&N/A	&N/A	&1.76$_{\pm0.00}$	&12.93$_{\pm0.00}$	&0.72$_{\pm0.46}$	&18.37$_{\pm0.00}$	&18.41$_{\pm0.00}$	&16.46$_{\pm0.00}$	&\textbf{19.60}$_{\pm0.00}$	\\
CIFAR-100	&N/A	&N/A	&N/A	&N/A	&N/A	&12.25$_{\pm0.00}$	&\textbf{14.89}$_{\pm0.00}$	&1.52$_{\pm1.14}$	&13.93$_{\pm0.00}$	&13.75$_{\pm0.00}$	&13.78$_{\pm0.00}$	&\textbf{14.89}$_{\pm0.00}$	\\
\midrule
Avg.score	&-	&-	&-	&-	&-	&28.95	&29.68	&18.50	&36.20	&35.05	&30.45	&\textbf{43.60}	\\
Avg.rank	&9.70	&7.90	&7.80	&7.70	&6.70	&6.30	&4.60	&8.90	&3.80	&3.70	&5.70	&\textbf{1.10}	\\
			\bottomrule
		\end{tabular}
}

	\begin{tablenotes}
		\footnotesize
		\item[1] * Note that N/A indicates the out-of-memory error, and the symbol ``-'' indicates that the average score cannot be calculated.
	\end{tablenotes}

\end{table*}

\begin{table*}[!th]
	\centering
	\newcommand{\tabincell}[2]{\begin{tabular}{@{}#1@{}}#2\end{tabular}}
	\caption{Average ACC(\%) over 20 runs by different multi-view clustering methods. The best result is highlighted in \textbf{bold}.}
	\label{tab:ACC}
	\resizebox{\textwidth}{!}{ \begin{tabular}{m{1.7cm}<{\centering}m{1.2cm}<{\centering}m{1.2cm}<{\centering}m{1.2cm}<{\centering}m{1.2cm}<{\centering}m{1.2cm}<{\centering}m{1.2cm}<{\centering}m{1.2cm}<{\centering}m{1.2cm}<{\centering}m{1.2cm}<{\centering}m{1.2cm}<{\centering}m{1.2cm}<{\centering}m{1.2cm}<{\centering}}
\toprule
Datasets	&MVSC	&AMGL	&MLAN	&SwMC	&{MVCTM}	&SFMC	&LMVSC	&{MSGL}	&SMVSC	&FPMVS-CAG	&{EOMSC-CA}	&\textbf{UDBGL}	\\
\midrule
WebKB-Texas	&55.13$_{\pm0.30}$	&51.79$_{\pm5.36}$	&49.63$_{\pm0.59}$	&54.55$_{\pm0.00}$	&63.64$_{\pm0.00}$	&40.11$_{\pm0.00}$	&60.96$_{\pm0.00}$	&45.40$_{\pm9.58}$	&56.68$_{\pm0.00}$	&57.75$_{\pm0.00}$	&57.22$_{\pm0.00}$	&\textbf{69.52}$_{\pm0.00}$	\\
MSRCv1	&56.98$_{\pm5.33}$	&64.90$_{\pm8.10}$	&73.81$_{\pm0.00}$	&70.48$_{\pm0.00}$	&\textbf{77.62}$_{\pm0.00}$	&56.67$_{\pm0.00}$	&34.29$_{\pm0.00}$	&39.31$_{\pm9.57}$	&67.14$_{\pm0.00}$	&60.00$_{\pm0.00}$	&51.90$_{\pm0.00}$	&74.76$_{\pm0.00}$	\\
Out-Scene	&31.82$_{\pm10.36}$	&49.91$_{\pm5.12}$	&40.55$_{\pm0.00}$	&45.46$_{\pm0.00}$	&50.97$_{\pm0.00}$	&47.84$_{\pm0.00}$	&59.86$_{\pm0.00}$	&26.69$_{\pm6.22}$	&63.32$_{\pm0.00}$	&63.58$_{\pm0.00}$	&52.27$_{\pm0.00}$	&\textbf{67.78}$_{\pm0.00}$	\\
Cora	&30.19$_{\pm0.11}$	&33.58$_{\pm3.00}$	&31.39$_{\pm0.00}$	&34.46$_{\pm0.13}$	&31.57$_{\pm0.00}$	&28.32$_{\pm0.00}$	&33.79$_{\pm0.00}$	&27.17$_{\pm4.29}$	&\textbf{53.29}$_{\pm0.00}$	&44.53$_{\pm0.00}$	&28.88$_{\pm0.00}$	&42.80$_{\pm0.00}$	\\
Citeseer	&21.24$_{\pm0.17}$	&21.48$_{\pm0.06}$	&22.64$_{\pm4.29}$	&21.62$_{\pm0.00}$	&22.07$_{\pm0.00}$	&24.67$_{\pm0.00}$	&29.35$_{\pm0.00}$	&22.18$_{\pm1.98}$	&41.94$_{\pm0.00}$	&37.53$_{\pm0.00}$	&32.00$_{\pm0.00}$	&\textbf{49.03}$_{\pm0.00}$	\\
{Notting-Hill}	&62.93$_{\pm14.53}$	&36.29$_{\pm1.84}$	&59.21$_{\pm0.00}$	&33.88$_{\pm0.00}$	&41.52$_{\pm0.00}$	&\textbf{91.89}$_{\pm0.00}$	&73.76$_{\pm0.00}$	&69.81$_{\pm14.87}$	&71.61$_{\pm0.00}$	&71.31$_{\pm0.00}$	&81.97$_{\pm0.00}$	&80.11$_{\pm0.00}$	\\
VGGFace	&N/A	&N/A	&N/A	&N/A	&N/A	&8.11$_{\pm0.00}$	&11.02$_{\pm0.00}$	&4.26$_{\pm1.75}$	&9.83$_{\pm0.00}$	&10.74$_{\pm0.00}$	&10.32$_{\pm0.00}$	&\textbf{12.31}$_{\pm0.00}$	\\
{YoutubeFaces}	&N/A	&N/A	&N/A	&N/A	&N/A	&71.64$_{\pm0.00}$	&64.93$_{\pm0.00}$	&60.87$_{\pm3.41}$	&66.18$_{\pm0.00}$	&62.02$_{\pm0.00}$	&60.53$_{\pm0.00}$	&\textbf{71.92}$_{\pm0.00}$	\\
CIFAR-10	&N/A	&N/A	&N/A	&N/A	&N/A	&11.52$_{\pm0.00}$	&25.01$_{\pm0.00}$	&11.76$_{\pm0.67}$	&29.20$_{\pm0.00}$	&29.55$_{\pm0.00}$	&28.25$_{\pm0.00}$	&\textbf{29.84}$_{\pm0.00}$	\\
CIFAR-100	&N/A	&N/A	&N/A	&N/A	&N/A	&6.87$_{\pm0.00}$	&8.12$_{\pm0.00}$	&1.82$_{\pm0.48}$	&8.14$_{\pm0.00}$	&\textbf{9.06}$_{\pm0.00}$	&8.28$_{\pm0.00}$	&\textbf{9.06}$_{\pm0.00}$	\\
\midrule
Avg.score	&-	&-	&-	&-	&-	&38.76	&40.11	&30.93	&46.73	&44.61	&41.16	&\textbf{50.71}	\\
Avg.rank	&8.70	&8.20	&7.90	&7.90	&6.70	&6.80	&4.90	&8.70	&3.70	&3.50	&5.40	&\textbf{1.50}	\\
		\bottomrule
	\end{tabular}
}
\end{table*}

\begin{table*}[!th]
	\centering
	\newcommand{\tabincell}[2]{\begin{tabular}{@{}#1@{}}#2\end{tabular}}
	\caption{Average PUR(\%) over 20 runs by different multi-view clustering methods. The best result is highlighted in \textbf{bold}.}
	\label{tab:PUR}
	\resizebox{\textwidth}{!}{ \begin{tabular}{m{1.7cm}<{\centering}m{1.2cm}<{\centering}m{1.2cm}<{\centering}m{1.2cm}<{\centering}m{1.2cm}<{\centering}m{1.2cm}<{\centering}m{1.2cm}<{\centering}m{1.2cm}<{\centering}m{1.2cm}<{\centering}m{1.2cm}<{\centering}m{1.2cm}<{\centering}m{1.2cm}<{\centering}m{1.2cm}<{\centering}}
\toprule
Datasets	&MVSC	&AMGL	&MLAN	&SwMC	&{MVCTM}	&SFMC	&LMVSC	&{MSGL}	&SMVSC	&FPMVS-CAG	&{EOMSC-CA}	&\textbf{UDBGL}	\\
\midrule
WebKB-Texas	&56.10$_{\pm0.16}$	&60.00$_{\pm1.98}$	&56.58$_{\pm0.88}$	&59.89$_{\pm0.00}$	&70.05$_{\pm0.00}$	&55.61$_{\pm0.00}$	&62.57$_{\pm0.00}$	&56.36$_{\pm2.41}$	&67.91$_{\pm0.00}$	&65.78$_{\pm0.00}$	&59.89$_{\pm0.00}$	&\textbf{72.73}$_{\pm0.00}$	\\
MSRCv1	&59.31$_{\pm5.06}$	&68.10$_{\pm7.17}$	&\textbf{80.48}$_{\pm0.00}$	&73.33$_{\pm0.00}$	&77.62$_{\pm0.00}$	&59.52$_{\pm0.00}$	&40.00$_{\pm0.00}$	&40.95$_{\pm9.27}$	&69.05$_{\pm0.00}$	&62.86$_{\pm0.00}$	&56.67$_{\pm0.00}$	&\textbf{80.48}$_{\pm0.00}$	\\
Out-Scene	&32.03$_{\pm10.40}$	&50.47$_{\pm5.32}$	&40.85$_{\pm0.00}$	&45.57$_{\pm0.00}$	&51.04$_{\pm0.00}$	&48.21$_{\pm0.00}$	&59.86$_{\pm0.00}$	&28.06$_{\pm6.47}$	&66.00$_{\pm0.00}$	&67.08$_{\pm0.00}$	&52.27$_{\pm0.00}$	&\textbf{67.78}$_{\pm0.00}$	\\
Cora	&30.34$_{\pm0.08}$	&35.54$_{\pm3.03}$	&31.79$_{\pm0.00}$	&36.70$_{\pm0.09}$	&33.64$_{\pm0.00}$	&32.90$_{\pm0.00}$	&38.88$_{\pm0.00}$	&30.43$_{\pm0.36}$	&53.66$_{\pm0.00}$	&45.49$_{\pm0.00}$	&34.34$_{\pm0.00}$	&\textbf{53.84}$_{\pm0.00}$	\\
Citeseer	&21.39$_{\pm0.11}$	&21.57$_{\pm0.05}$	&22.80$_{\pm4.39}$	&22.04$_{\pm0.00}$	&22.61$_{\pm0.00}$	&25.30$_{\pm0.00}$	&31.76$_{\pm0.00}$	&22.32$_{\pm2.09}$	&47.01$_{\pm0.00}$	&42.81$_{\pm0.00}$	&45.35$_{\pm0.00}$	&\textbf{51.15}$_{\pm0.00}$	\\
{Notting-Hill}	&65.24$_{\pm13.91}$	&37.07$_{\pm2.10}$	&\textbf{92.85}$_{\pm0.00}$	&37.10$_{\pm0.00}$	&71.63$_{\pm0.00}$	&91.89$_{\pm0.00}$	&81.20$_{\pm0.00}$	&75.37$_{\pm13.29}$	&78.41$_{\pm0.00}$	&78.50$_{\pm0.00}$	&84.70$_{\pm0.00}$	&87.23$_{\pm0.00}$	\\
VGGFace	&N/A	&N/A	&N/A	&N/A	&N/A	&8.52$_{\pm0.00}$	&11.84$_{\pm0.00}$	&4.43$_{\pm1.85}$	&10.28$_{\pm0.00}$	&11.15$_{\pm0.00}$	&11.17$_{\pm0.00}$	&\textbf{13.29}$_{\pm0.00}$	\\
{YoutubeFaces}	&N/A	&N/A	&N/A	&N/A	&N/A	&78.94$_{\pm0.00}$	&72.02$_{\pm0.00}$	&66.18$_{\pm2.50}$	&69.99$_{\pm0.00}$	&71.90$_{\pm0.00}$	&62.61$_{\pm0.00}$	&\textbf{79.48}$_{\pm0.00}$	\\
CIFAR-10	&N/A	&N/A	&N/A	&N/A	&N/A	&11.92$_{\pm0.00}$	&27.65$_{\pm0.00}$	&11.91$_{\pm0.75}$	&32.40$_{\pm0.00}$	&\textbf{32.85}$_{\pm0.00}$	&30.57$_{\pm0.00}$	&31.61$_{\pm0.00}$	\\
CIFAR-100	&N/A	&N/A	&N/A	&N/A	&N/A	&7.57$_{\pm0.00}$	&9.66$_{\pm0.00}$	&1.89$_{\pm0.53}$	&8.64$_{\pm0.00}$	&9.45$_{\pm0.00}$	&8.86$_{\pm0.00}$	&\textbf{9.83}$_{\pm0.00}$	\\
\midrule
Avg.score	&-	&-	&-	&-	&-	&42.04	&43.54	&33.79	&50.33	&48.79	&44.64	&\textbf{54.74}	\\
Avg.rank	&9.70	&8.00	&7.00	&7.80	&6.80	&6.50	&4.70	&8.80	&3.90	&3.80	&5.40	&\textbf{1.40}	\\
		\bottomrule
	\end{tabular}
}
\end{table*}

\subsection{Datasets and Evaluation Metrics}
In the experimental evaluation, ten real-world multi-view datasets are used, namely, WebKB-Texas \cite{Cai2023}, MSRCv1 \cite{ChenManSheng}, Out-Scene \cite{D-Out-Scene}, Cora \cite{D-Citeseer}, Citeseer \cite{D-Citeseer}, Notting-Hill, VGGFace\footnotemark[1], YoutubeFaces, CIFAR-10, and CIFAR-100. Note that CIFAR-10 and CIFAR-100 are two versions of the CIFAR\footnotemark[2] dataset. The details of these datasets are given in Table~\ref{tab:datasets}.

\footnotetext[1]{\url{https://www.robots.ox.ac.uk/~vgg/data/vgg_face2/}}
\footnotetext[2]{\url{https://www.cs.toronto.edu/~kriz/cifar.html}}

The clustering results are evaluated by three widely-used metrics, i.e., the normalized mutual information (NMI) \cite{Huang2021}, the accuracy (ACC) \cite{Deng23_SACC}, and the purity (PUR) \cite{LMVSC}. For all the three evaluation metrics, larger values indicate better clustering performance.

\subsection{Baseline Methods and Experimental Settings}

We experimentally compare the proposed UDBGL method against the eleven baseline MVC methods, namely
	multi-view spectral clustering (MVSC) \cite{MVSC}
	auto-weighted multiple graph learning (AMGL) \cite{AMGL},
	multi-view clustering with adaptive neighbors (MLAN) \cite{MLAN},
	self-weighted multiview clustering (SwMC) \cite{SwMC},
    multi-view clustering on topological manifold (MVCTM) \cite{MVCTM},
	scalableand parameter-free multi-view graph clustering (SFMC) \cite{SFMC},
	large-scale multi-view subspace clustering (LMVSC) \cite{LMVSC},
    multi-view structured graph learning (MSGL) \cite{MSGL},
	scalable multi-view subspace clustering (SMVSC) \cite{SMVSC},
	fast parameter-free multi-view subspace clustering with consensus anchor guidance (FPMVS-CAG) \cite{FPMVS-CAG},
    and efficient one-pass multi-view subspace clustering with consensus anchors (EOMSC-CA) \cite{EOMSC-CA}.

Among these baseline methods, AMGL, MLAN, SwMC, and MVCTM are four (general) graph learning based methods, while MVSC, SFMC, LMVSC, MSGL, SMVSC, FPMVS-CAG, and EOMSC-CA are seven bipartite graph-based methods. For the proposed method and the baseline methods, the experiments on each dataset are repeated $20$ times, and their average scores and standard deviations will be reported. The hyper-parameters in the proposed method and the baseline methods are tuned in the range of $\{10^{-3},10^{-2},\cdots,10^3\}$, unless the value (or range) of the hyper-parameter is specified by the corresponding paper. The number of anchors is tuned in the range of $\{c,50,100,200\}$. To avoid the expensive computational costs of parameter-tuning on the whole large-scale datasets, for all the test methods, if $n>10,000$, then we will tune the hyper-parameters using a random subset of $10,000$ samples.

\subsection{Comparison with Other MVC Methods}

In this section, we compare the proposed UDBGL method against the eleven baseline methods on the benchmark datasets. Specifically, the clustering scores w.r.t. NMI, ACC, and PUR by different MVC methods are reported in Tables~\ref{tab:NMI}, \ref{tab:ACC}, and \ref{tab:PUR}, respectively. The best score on each dataset is highlighted in bold, and ``N/A'' indicates that this method is not computationally feasible on this dataset due to the out-of-memory error.
As shown in Table~\ref{tab:NMI}, in terms of NMI, our UDBGL method outperforms all the baseline methods on nine out of the ten datasets.  Particularly, in comparison with the general graph-based methods (i.e., AMGL, MLAN, SwMC, and MVCTM), the bipartite graph-based methods can generally handle the large datasets more efficiently and produce more robust clustering results in most cases. Among the bipartite graph-based methods (i.e., MVSC, SFMC, LMVSC, MSGL, SMVSC, FPMVS-CAG, EOMSC-CA, and UDBGL), the proposed UDBGL method is able to produce consistently better clustering results on most of the datasets.
Although the FPMVS-CAG method yields a higher NMI score than UDBGL on the Out-Scene dataset, yet the UDBGL method outperforms FPMVS-CAG on all the other datasets.
Similar advantages of UDBGL w.r.t. ACC and PUR can also be observed in Tables~\ref{tab:ACC} and \ref{tab:PUR}, respectively.

\begin{figure}[!t] 
	\begin{center}
		{\subfigure[{\scriptsize WebKB-Texas}]
			{\includegraphics[width=0.22\columnwidth]{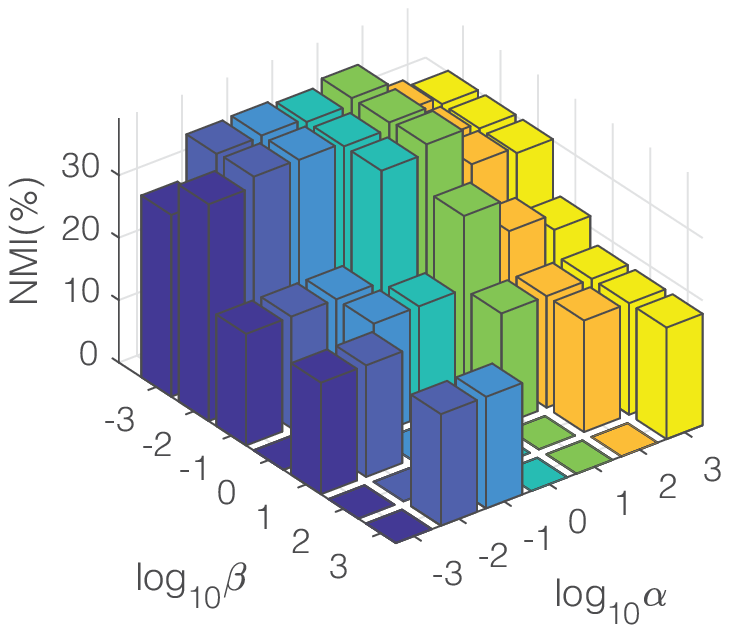}}}
		{\subfigure[{\scriptsize MSRC-v1}]
			{\includegraphics[width=0.22\columnwidth]{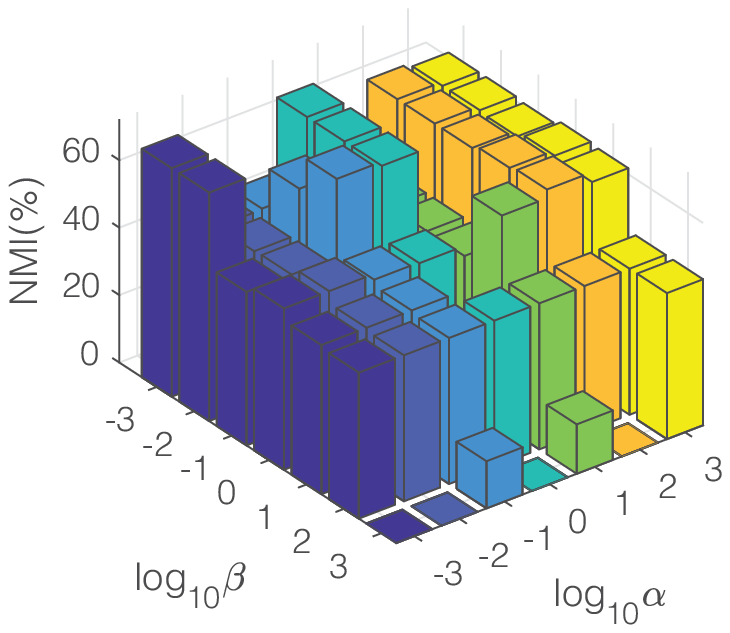}}}
		{\subfigure[{\scriptsize Out-Scene}]
			{\includegraphics[width=0.22\columnwidth]{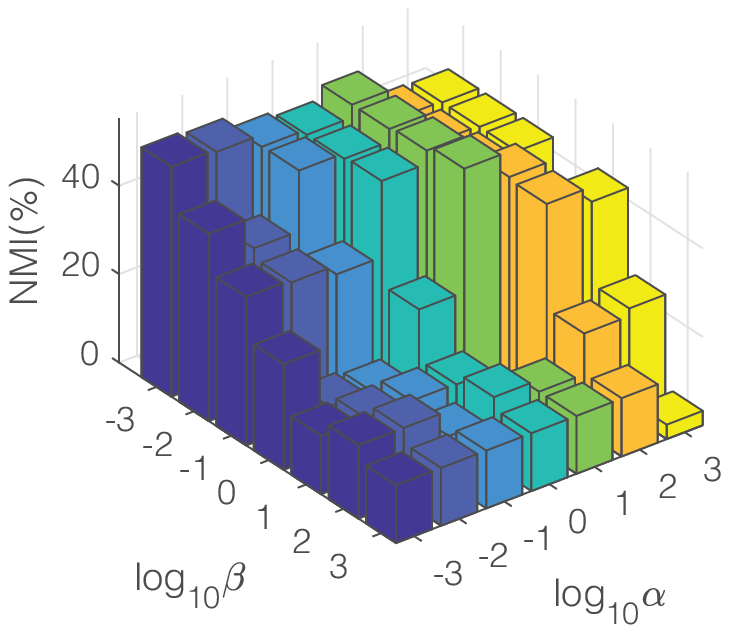}}}
		{\subfigure[{\scriptsize Cora}]
			{\includegraphics[width=0.22\columnwidth]{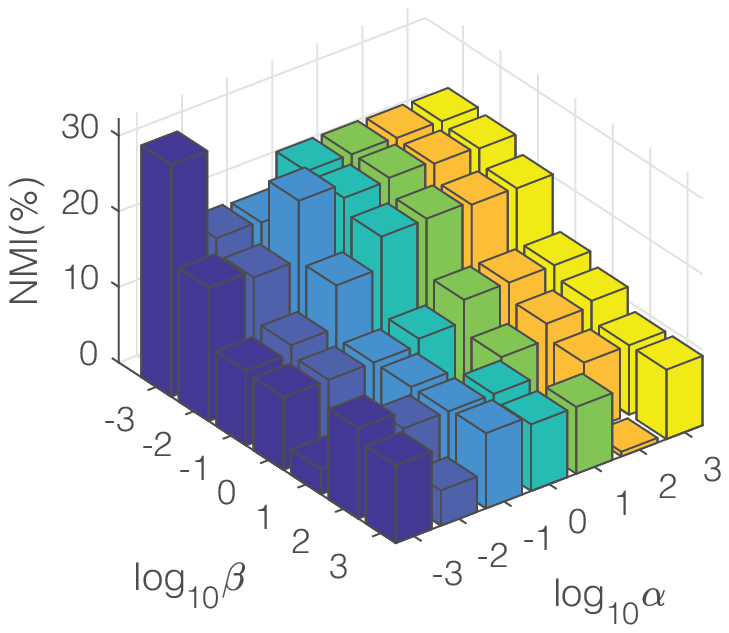}}}\vskip 0.008in
		{\subfigure[{\scriptsize Citeseer}]
			{\includegraphics[width=0.22\columnwidth]{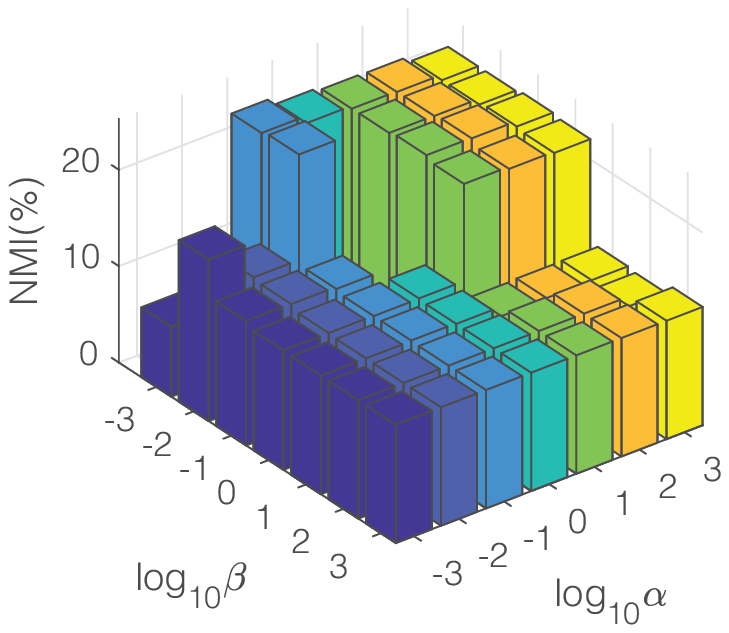}}}
		{\subfigure[{\scriptsize Notting-Hill}]
			{\includegraphics[width=0.22\columnwidth]{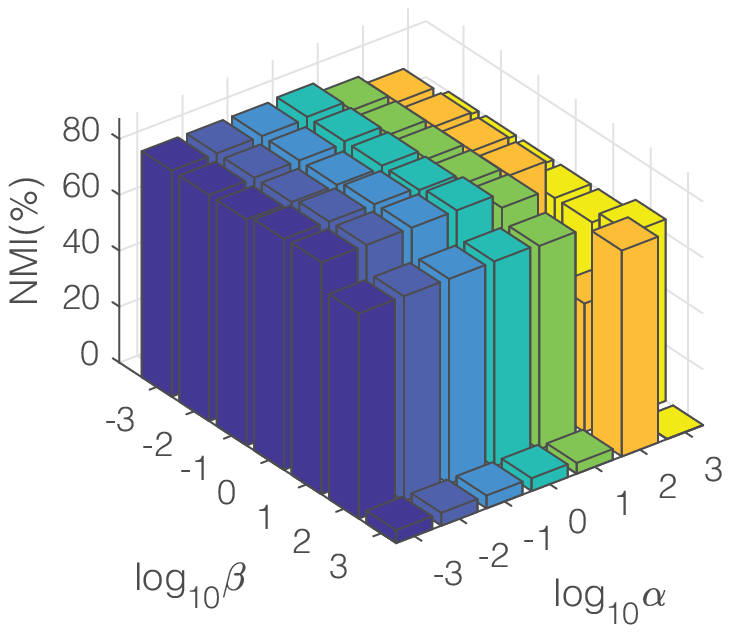}}}
		{\subfigure[{\scriptsize VGGFace}]
			{\includegraphics[width=0.22\columnwidth]{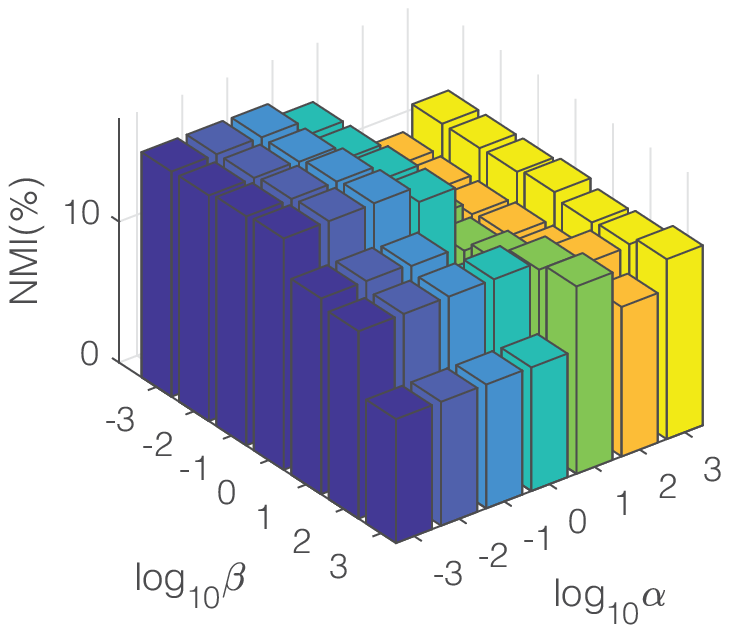}}}
		{\subfigure[{\scriptsize YoutubeFaces}]
			{\includegraphics[width=0.22\columnwidth]{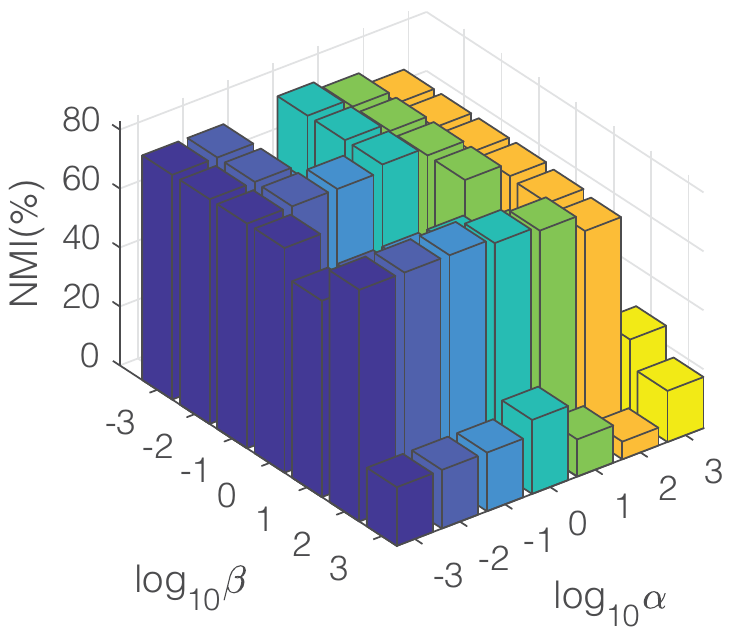}}}\vskip 0.008in
		{\subfigure[{\scriptsize CIFAR-10}]
			{\includegraphics[width=0.22\columnwidth]{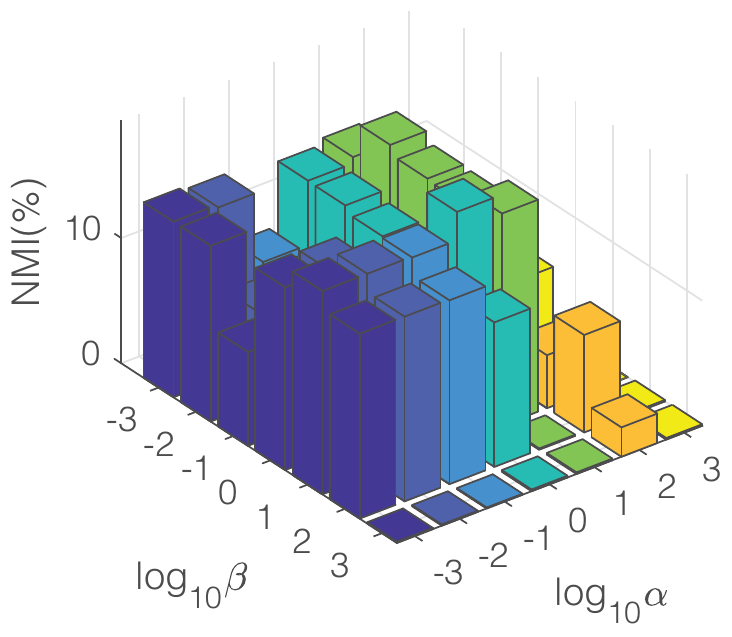}}}
		{\subfigure[{\scriptsize CIFAR-100}]
			{\includegraphics[width=0.22\columnwidth]{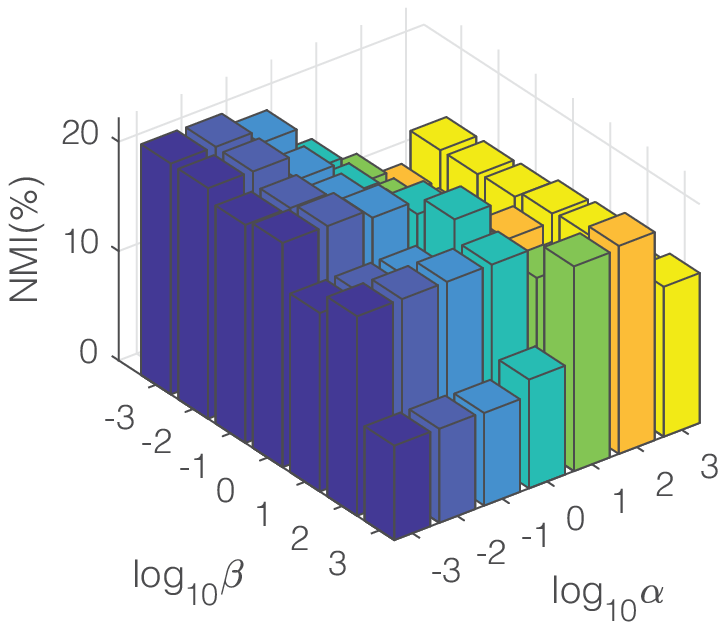}}}\vskip -0.09in
		\caption{The clustering performance (w.r.t. NMI) by UDBGL with varying values of parameters $\alpha$ and $\beta$.}
		\label{fig:sensitivity}
	\end{center}
\end{figure}

Further, we report the average scores and the average ranks (across all benchmark datasets) by different MVC methods in Tables~\ref{tab:NMI}, \ref{tab:ACC}, and \ref{tab:PUR}. Note that, among the twelve MVC methods, if only seven MVC methods are computational feasible and the other five methods are infeasible due to the the out-of-memory error on a dataset, then the five infeasible methods will be equally ranked in the eighth position on this dataset. In terms of the average score, UDBGL achieves the average NMI(\%), ACC(\%), and PUR(\%) scores of 43.60, 50.71, 54.74, respectively, which substantially outperforms the second best average NMI(\%), ACC(\%), and PUR(\%) scores of 36.20, 46.73, and 50.33, respectively. In terms of the average rank, UDBGL obtains the average ranks (w.r.t. NMI, ACC, and PUR) of 1.10, 1.50, and 1.40, respectively, while the second best method only obtains the average ranks of 3.70, 3.50, and 3.80, respectively, which demonstrate the robustness of the proposed UDBGL method over the other MVC methods.

\begin{figure}[!t] 
	\begin{center}
		{\subfigure[{\scriptsize WebKB-Texas}]
			{\includegraphics[width=0.22\columnwidth]{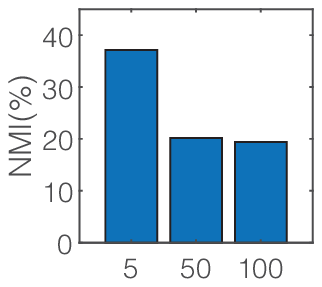}}}
		{\subfigure[{\scriptsize MSRC-v1}]
			{\includegraphics[width=0.22\columnwidth]{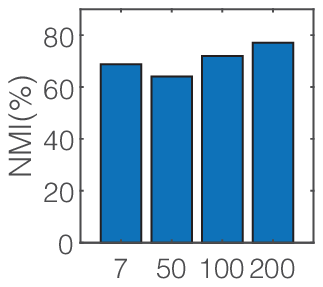}}}
		{\subfigure[{\scriptsize Out-Scene}]
			{\includegraphics[width=0.22\columnwidth]{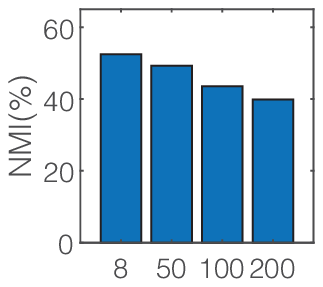}}}
		{\subfigure[{\scriptsize Cora}]
			{\includegraphics[width=0.22\columnwidth]{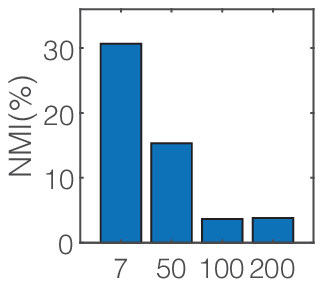}}}\vskip 0.008in
		{\subfigure[{\scriptsize Citeseer}]
			{\includegraphics[width=0.22\columnwidth]{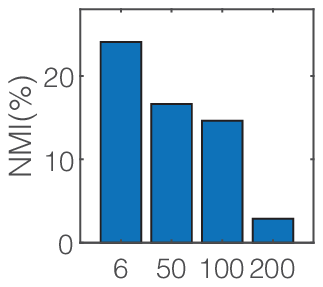}}}
        {\subfigure[{\scriptsize Notting-Hill}]
			{\includegraphics[width=0.22\columnwidth]{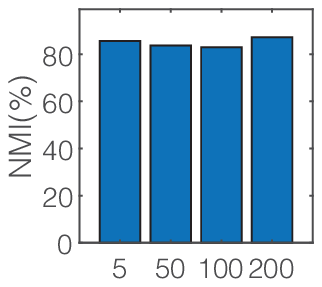}}}
		{\subfigure[{\scriptsize VGGFace}]
			{\includegraphics[width=0.22\columnwidth]{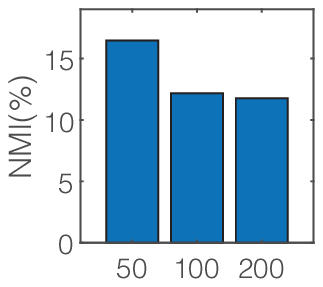}}}
        {\subfigure[{\scriptsize YoutubeFaces}]
			{\includegraphics[width=0.22\columnwidth]{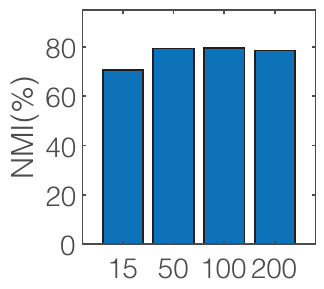}}}\vskip 0.008in
		{\subfigure[{\scriptsize CIFAR-10}]
			{\includegraphics[width=0.22\columnwidth]{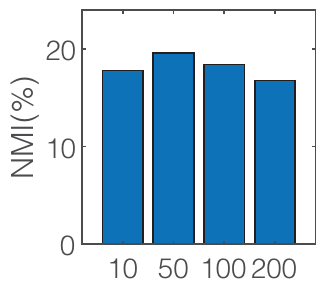}}}
		{\subfigure[{\scriptsize CIFAR-100}]
			{\includegraphics[width=0.22\columnwidth]{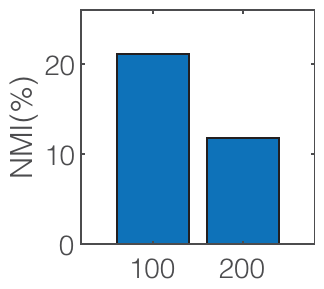}\label{fig:para_anchors_j}}}\vskip -0.09in
		\caption{The clustering performance (w.r.t. NMI) by UDBGL with varying number of anchors.}
		\label{fig:para_anchors}
	\end{center}
\end{figure}

\subsection{Parameter Analysis}
In this section, we test the influence of the hyper-parameters $\alpha$ and $\beta$ as well as the number of anchors $m$ in the proposed UDBGL method.

In UDBGL, the two hyper-parameters, i.e., $\alpha$ and $\beta$, are utilized to negotiate the importance of the regularization term (of $\textbf{Z}^{(v)}$) and the graph fusion term. The clustering performance of UDBGL with varying hyper-parameters $\alpha$ and $\beta$ is illustrated in Fig. \ref{fig:sensitivity}, which shows that moderate parameter values are generally beneficial to the clustering performance on various datasets.

Then, we test the clustering performance of the proposed UDBGL method with the number of anchors varying in the range of $\{c,50,100,200\}$, which corresponds to the four bars in each sub-figure of Fig.~\ref{fig:para_anchors}. Since the anchor number should be greater than or equal to the number of clusters (i.e., $c$), the number of tested values may be less than four on some datasets, e.g., on the CIFAR-100 dataset (with $c=100$) only two values (i.e., 100 and 200) are tested, corresponding to the two bars in Fig.~\ref{fig:para_anchors_j}. As can be observed in Fig.~3, setting the anchor number equal to the cluster number $c$ can lead to quite robust clustering results on most of datasets.

\begin{figure}[!t] 
	\begin{center}
		{\subfigure[{\scriptsize WebKB-Texas}]
			{\includegraphics[width=0.22\columnwidth]{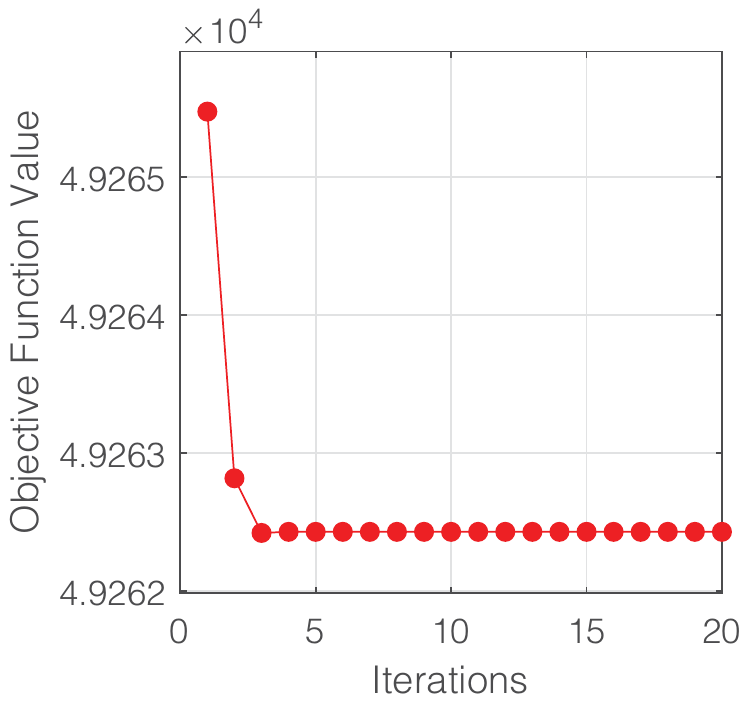}}}
		{\subfigure[{\scriptsize MSRC-v1}]
			{\includegraphics[width=0.22\columnwidth]{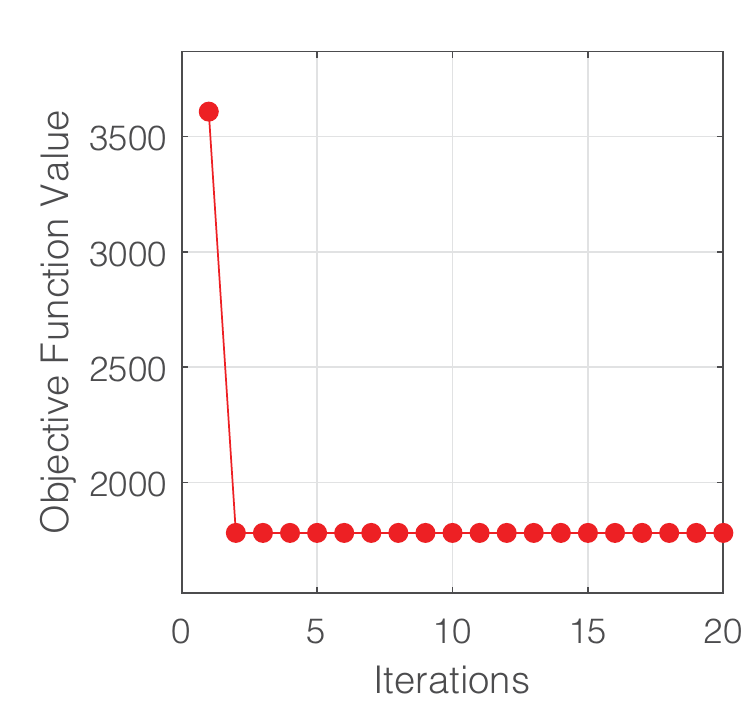}}}
		{\subfigure[{\scriptsize Out-Scene}]
			{\includegraphics[width=0.22\columnwidth]{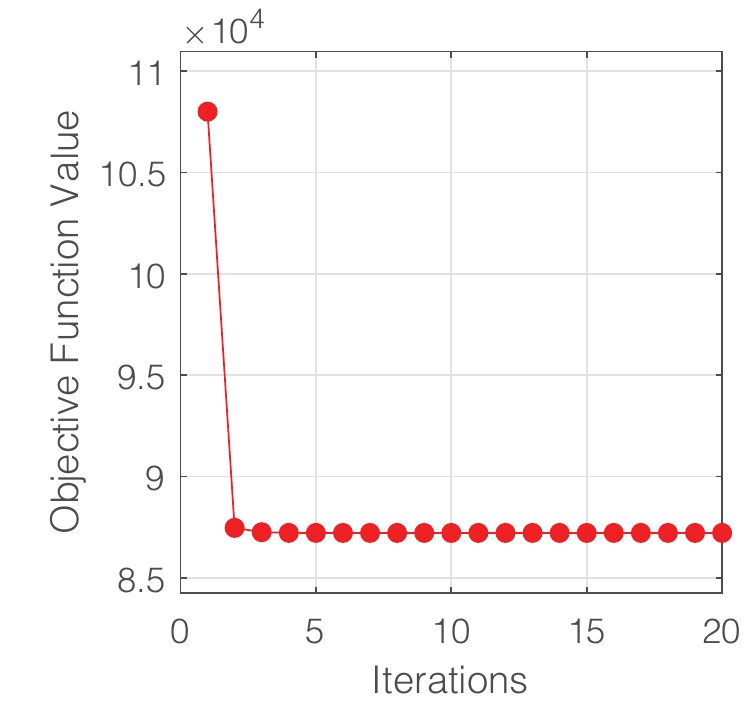}}}
		{\subfigure[{\scriptsize Cora}]
			{\includegraphics[width=0.22\columnwidth]{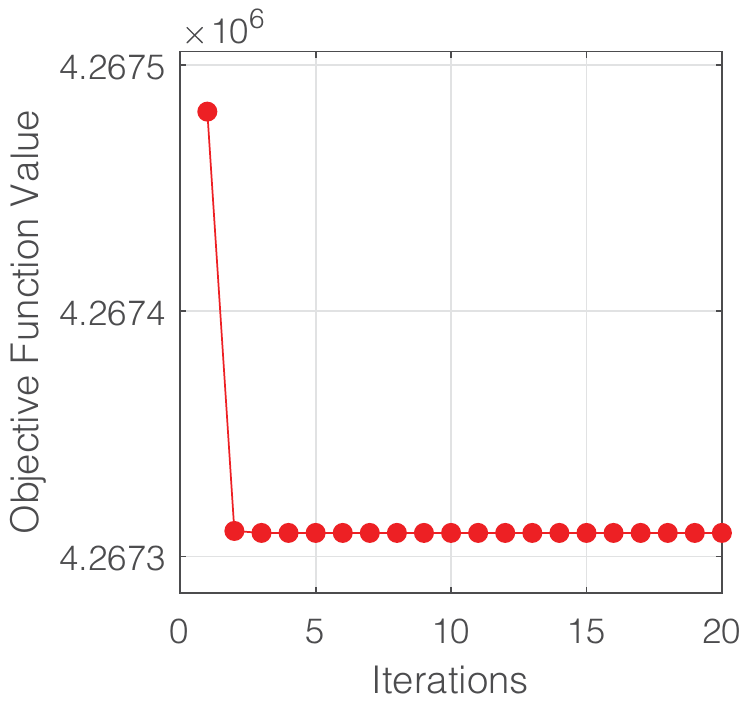}}}\vskip 0.008in
		{\subfigure[{\scriptsize Citeseer}]
			{\includegraphics[width=0.22\columnwidth]{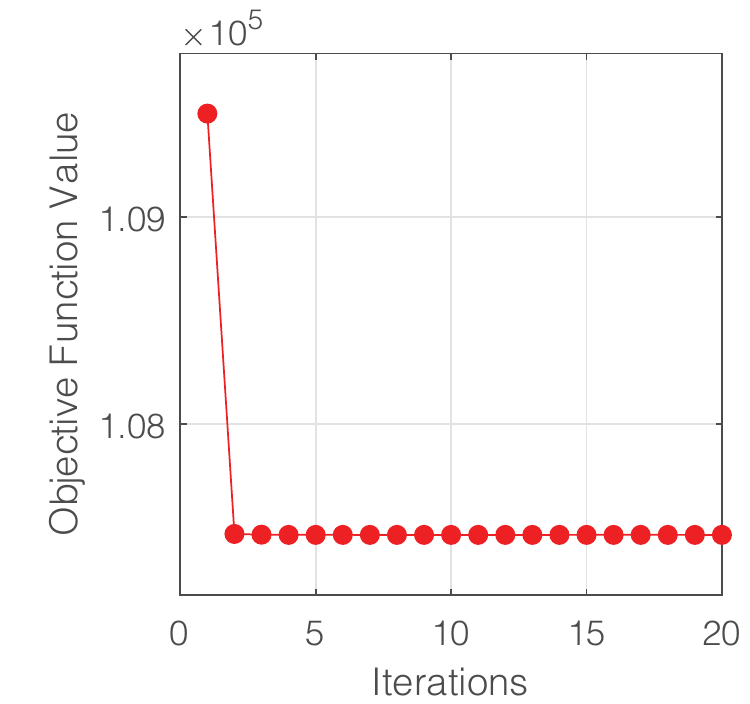}}}
		{\subfigure[{\scriptsize Notting-Hill}]
			{\includegraphics[width=0.22\columnwidth]{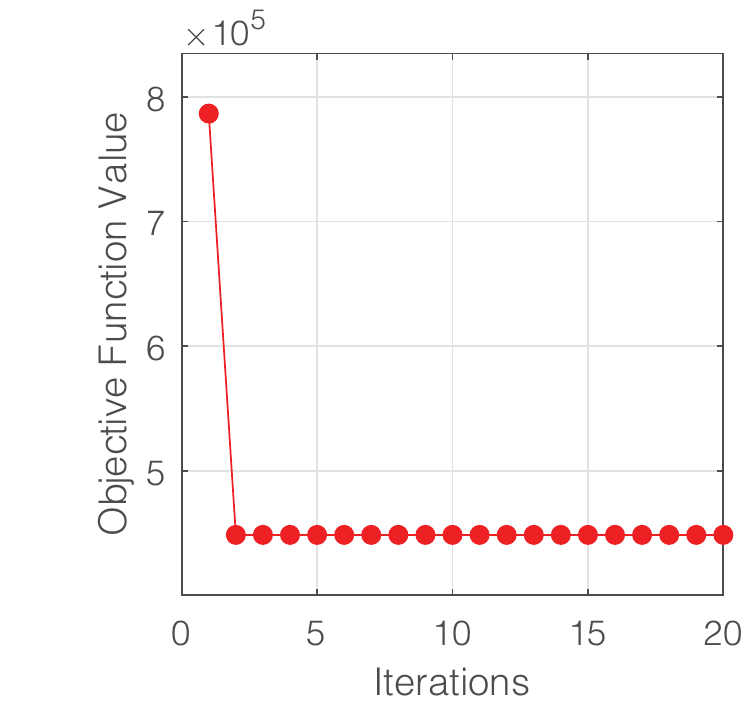}}}
		{\subfigure[{\scriptsize VGGFace}]
			{\includegraphics[width=0.22\columnwidth]{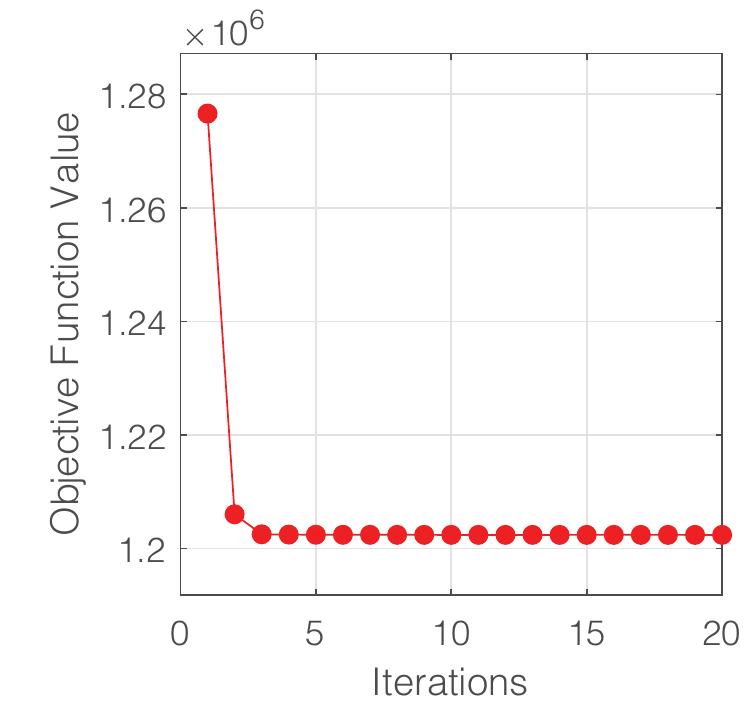}}}
		{\subfigure[{\scriptsize YoutubeFaces}]
			{\includegraphics[width=0.22\columnwidth]{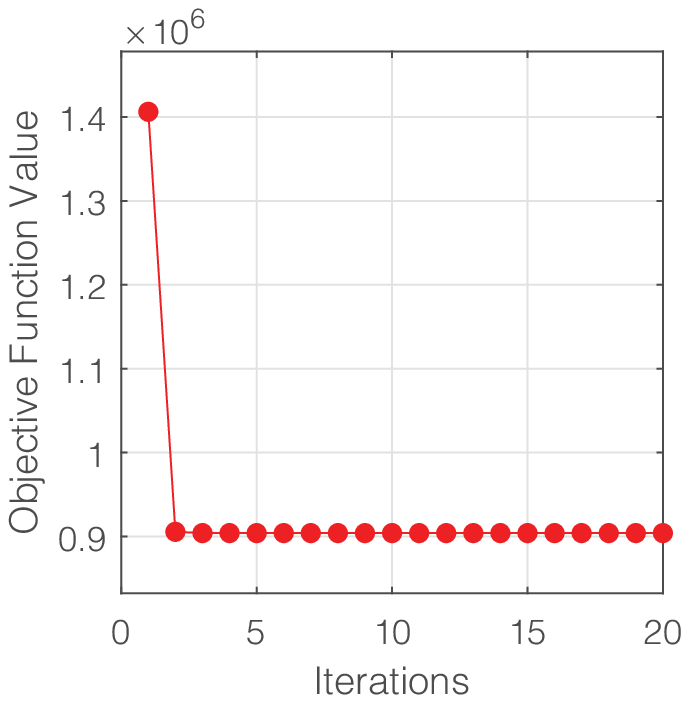}}}\vskip 0.008in
		{\subfigure[{\scriptsize CIFAR-10}]
			{\includegraphics[width=0.22\columnwidth]{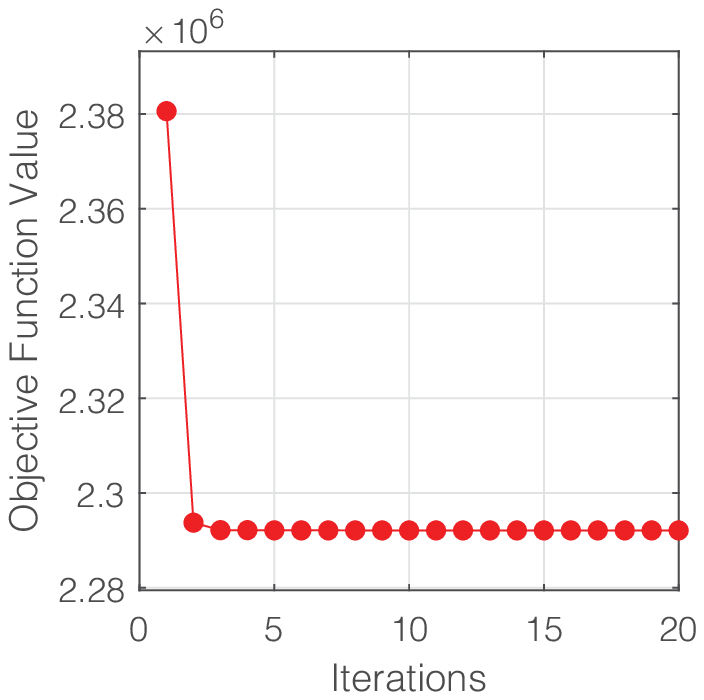}}}
		{\subfigure[{\scriptsize CIFAR-100}]
			{\includegraphics[width=0.22\columnwidth]{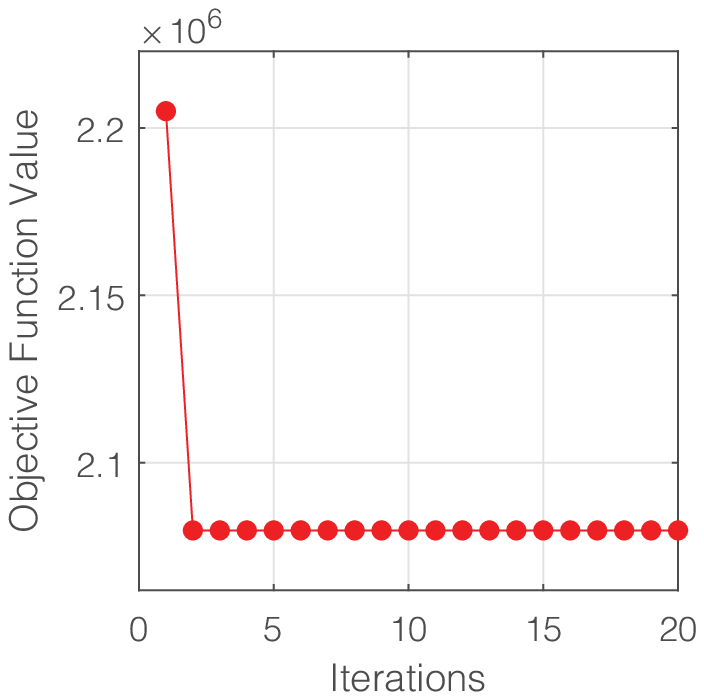}}}
		\caption{Convergence of the objective value of UDBGL with increasing iterations.}\vskip -0.09in
		\label{fig:convergence}
	\end{center}
\end{figure}

\subsection{Empirical Convergence Analysis}
\label{sec:empirical_convergence}
In this section, we test the convergence of UDBGL on the benchmark datasets.
Fig. \ref{fig:convergence} shows the convergence curves of the objective function values of UDBGL after different iterations. As can be seen in Fig. \ref{fig:convergence}, the objective function value decreases very fast and generally converges within twenty iterations on various datasets, which shows the good convergence property of the proposed UDBGL method.

\subsection{Ablation Study}

\begin{figure}[!t] 
	\begin{center}
		{\subfigure[{\scriptsize WebKB-Texas}]
			{\includegraphics[width=0.22\columnwidth]{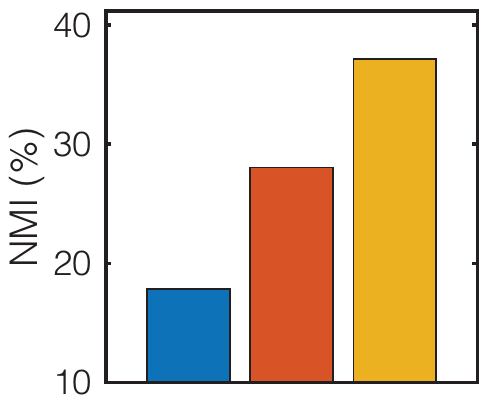}}}
		{\subfigure[{\scriptsize MSRC-v1}]
			{\includegraphics[width=0.22\columnwidth]{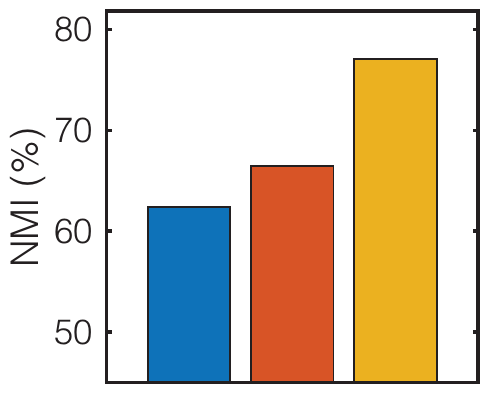}}}
		{\subfigure[{\scriptsize Out-Scene}]
			{\includegraphics[width=0.22\columnwidth]{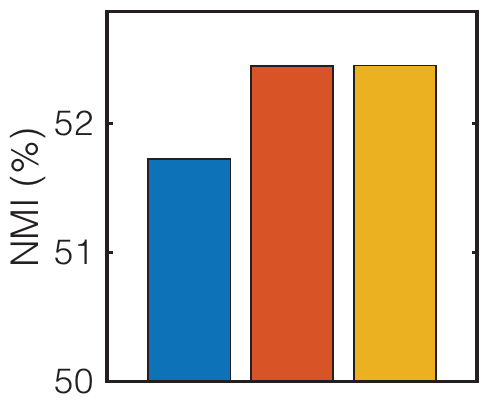}}}
		{\subfigure[{\scriptsize Cora}]
			{\includegraphics[width=0.22\columnwidth]{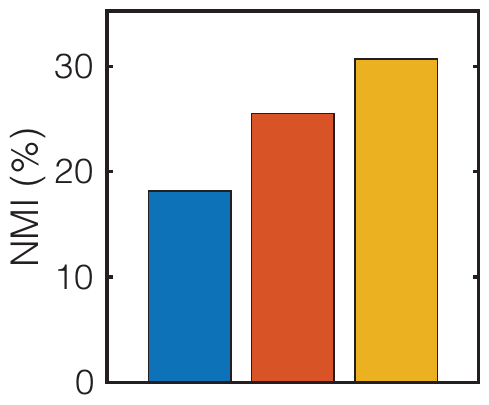}}}\vskip 0.008in
		{\subfigure[{\scriptsize Citeseer}]
			{\includegraphics[width=0.22\columnwidth]{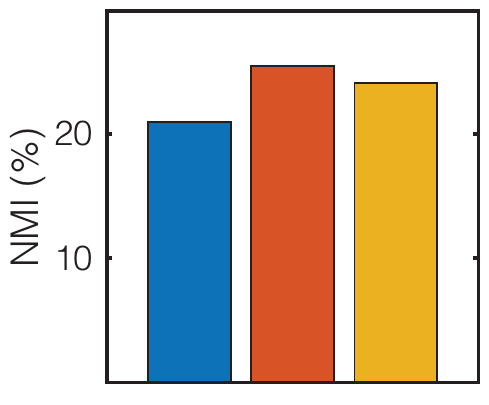}}}
		{\subfigure[{\scriptsize Notting-Hill}]
			{\includegraphics[width=0.22\columnwidth]{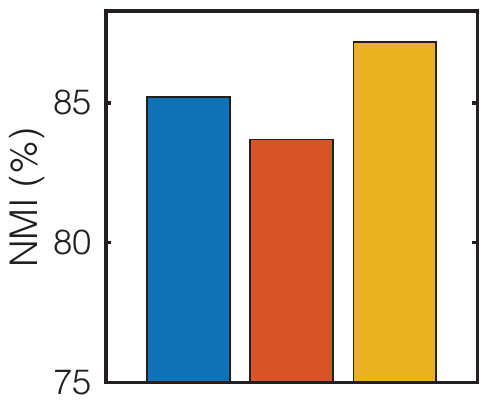}}}
		{\subfigure[{\scriptsize VGGFace}]
			{\includegraphics[width=0.22\columnwidth]{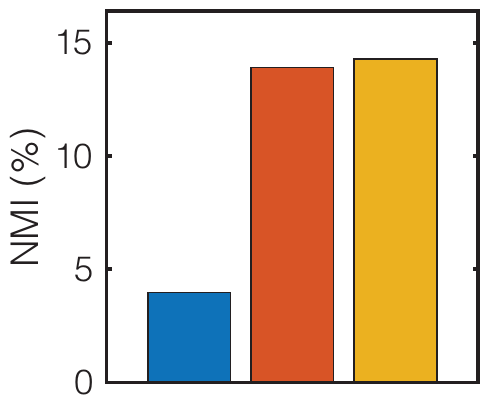}}}
		{\subfigure[{\scriptsize YoutubeFaces}]
			{\includegraphics[width=0.22\columnwidth]{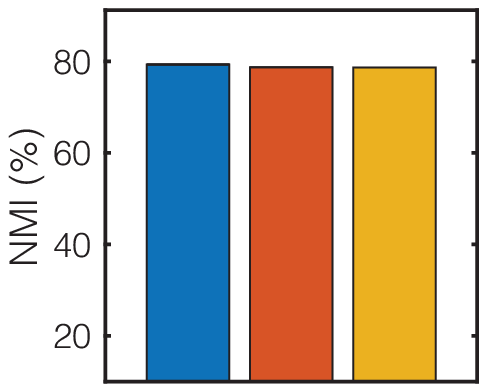}}}\vskip 0.008in
		{\subfigure[{\scriptsize CIFAR-10}]
			{\includegraphics[width=0.22\columnwidth]{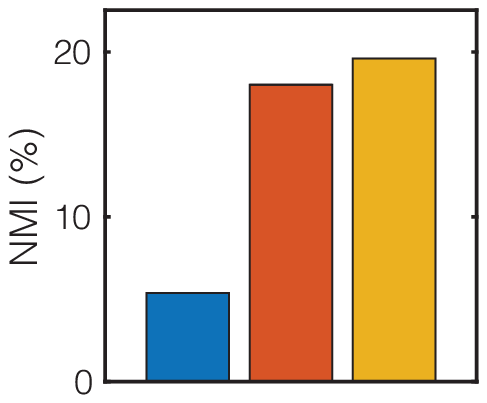}}}
		{\subfigure[{\scriptsize CIFAR-100}]
			{\includegraphics[width=0.22\columnwidth]{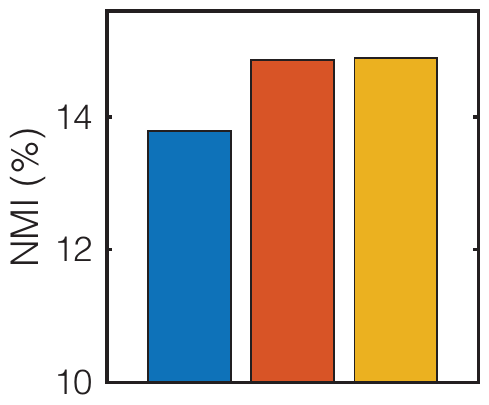}}}\vskip 0.008in
		{\subfigure
			{\includegraphics[width=0.97\columnwidth]{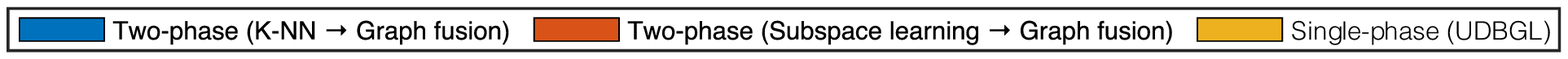}}}
		\caption{Ablation analysis on the benchmark datasets.}
		\label{fig:ablation}
	\end{center}
\end{figure}

In this section, we test the influence of the single-phase formulation of UDBGL by comparing it with the two-phase versions. Specifically, we design two variants by removing the view-specific graph learning part or by splitting the model into two phases in the ablation study. Our first variant is to first use the $K$-nearest neighbor ($K$-NN) graphs as the pre-constructed graphs and then perform graph fusion to learn a view-consensus graph with discrete cluster structure, which can be viewed as removing the view-specific graph learning part from UDBGL. The second variant is to treat the view-specific graph learning and the view-consensus graph learning in UDBGL as two separate phases, which first enforces subspace learning to obtain the view-specific bipartite graphs and then performs graph fusion to achieve the final clustering. As shown in Fig.~\ref{fig:ablation}, the simultaneous learning of the view-specific graphs and the view-consensus graph can generally lead to more robust clustering performance than the two variants across the ten benchmark datasets. In particular, the clustering performances of the two variants have shown instability across different datasets, which means that they perform well on some datasets but may significantly decline on some other datasets (e.g., on the WebKB-Texas, MSRC-v1, Cora, Notting-Hill, and CIFAR-10 datasets). In comparison with the two variants, the unified formulation of UDBGL achieves consistently high-quality clustering performance across the ten datasets, which is probably due to its ability to learn the view-specific graphs and the view-consensus graph jointly and adaptively.

\subsection{Execution Time}
\begin{figure*}[!th] 
	\centering
	\includegraphics[width=0.9\textwidth]{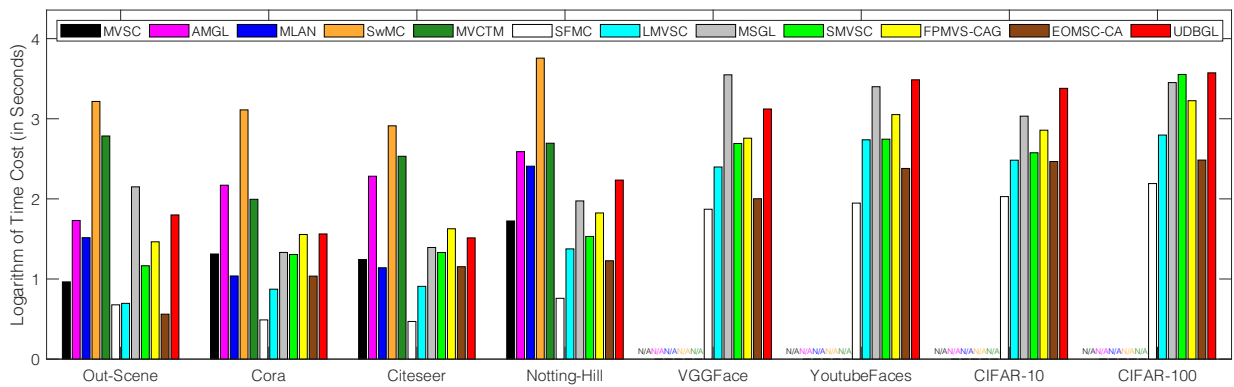}
	\caption{Execution times of different MVC methods on the eight larger datasets (with $n>2,000$). Note that N/A indicates the out-of-memory error.}
	\label{fig:time}
\end{figure*}

In this section, we report the execution times of our UDBGL method and the baseline methods on the eight larger datasets (with $n>2,000$).
As shown in Fig. \ref{fig:time}, the MVSC, AMGL, MLAN, SwMC, and MVCTM methods are not computationally feasible on the four large-scale datasets (with over $30,000$ samples), while the other seven bipartite graph-based methods are computationally feasible on all the datasets. Particularly, the proposed UDBGL method has shown comparable time cost to the MSGL, SMVSC, and FPMVS-CAG methods on most of the benchmark datasets. From the performance results in Tables~\ref{tab:NMI}, \ref{tab:ACC}, and \ref{tab:PUR} and the execution times in Fig. \ref{fig:time}, we can observe that our UDBGL method is able to achieve advantageous clustering performance on a variety of datasets while maintaining highly-competitive efficiency.

\section{Conclusion}
\label{sec:conclusion}
In this paper, we propose an efficient MVC approach termed UDBGL. In particular, the anchor-based subspace learning is enforced to learn the view-specific bipartite graphs from multiple views, which are fused into a view-consensus bipartite graph with adaptive view weights. Remarkably, the view-specific bipartite graph learning and the view-consensus bipartite graph learning can promote each other mutually and adaptively, which are further coupled with the Laplacian low rank constraint to directly obtain the discrete cluster structure from the fused graph. An alternating minimization algorithm is designed to optimize the proposed model, whose computational complexity is linear to the data size. Extensive experiments on ten multi-view datasets, whose data sizes range from $187$ to $60,000$, have demonstrated the clustering robustness and efficiency of our UDBGL approach over the state-of-the-art MVC approaches.

\section*{Acknowledgments}

This work was supported by the NSFC (61976097, 62276277, 62077045, \& U1811263), the Natural Science Foundation of Guangdong Province (2021A1515012203), and Guangdong Basic and Applied Basic Research Foundation (2022B1515120059).

\bibliographystyle{IEEEtran}
\bibliography{UDBGL-ref}

\begin{thebibliography}{10}
\providecommand{\url}[1]{#1}
\csname url@samestyle\endcsname
\providecommand{\newblock}{\relax}
\providecommand{\bibinfo}[2]{#2}
\providecommand{\BIBentrySTDinterwordspacing}{\spaceskip=0pt\relax}
\providecommand{\BIBentryALTinterwordstretchfactor}{4}
\providecommand{\BIBentryALTinterwordspacing}{\spaceskip=\fontdimen2\font plus
\BIBentryALTinterwordstretchfactor\fontdimen3\font minus
  \fontdimen4\font\relax}
\providecommand{\BIBforeignlanguage}[2]{{%
\expandafter\ifx\csname l@#1\endcsname\relax
\typeout{** WARNING: IEEEtran.bst: No hyphenation pattern has been}%
\typeout{** loaded for the language `#1'. Using the pattern for}%
\typeout{** the default language instead.}%
\else
\language=\csname l@#1\endcsname
\fi
#2}}
\providecommand{\BIBdecl}{\relax}
\BIBdecl

\bibitem{chao2021survey}
G.~Chao, S.~Sun, and J.~Bi, ``A survey on multiview clustering,'' \emph{IEEE
  Transactions on Artificial Intelligence}, vol.~2, no.~2, pp. 146--168, 2021.

\bibitem{SwMC}
F.~Nie, J.~Li, X.~Li \emph{et~al.}, ``Self-weighted multiview clustering with
  multiple graphs.'' in \emph{Proc. of International Joint Conference on
  Artificial Intelligence (IJCAI)}, 2017, pp. 2564--2570.

\bibitem{MCGC}
K.~Zhan, F.~Nie, J.~Wang, and Y.~Yang, ``Multiview consensus graph
  clustering,'' \emph{IEEE Transactions on Image Processing}, vol.~28, no.~3,
  pp. 1261--1270, 2018.

\bibitem{GMC}
H.~Wang, Y.~Yang, and B.~Liu, ``Gmc: Graph-based multi-view clustering,''
  \emph{IEEE Transactions on Knowledge and Data Engineering}, vol.~32, no.~6,
  pp. 1116--1129, 2019.

\bibitem{liangTNNLS}
Y.~Liang, D.~Huang, C.-D. Wang, and P.~S. Yu, ``Multi-view graph learning by
  joint modeling of consistency and inconsistency,'' \emph{IEEE Transactions on
  Neural Networks and Learning Systems, in press}, 2022.

\bibitem{RG-MVC}
W.~Liang, X.~Liu, S.~Zhou, J.~Liu, S.~Wang, and E.~Zhu, ``Robust graph-based
  multi-view clustering,'' in \emph{Proc. of AAAI Conference on Artificial
  Intelligence}, 2022, pp. 7462--7469.

\bibitem{AMGL}
F.~Nie, J.~Li, X.~Li \emph{et~al.}, ``Parameter-free auto-weighted multiple
  graph learning: a framework for multiview clustering and semi-supervised
  classification.'' in \emph{Proc. of International Joint Conference on
  Artificial Intelligence (IJCAI)}, 2016, pp. 1881--1887.

\bibitem{Li2022}
X.-L. Li, M.-S. Chen, C.-D. Wang, and J.-H. Lai, ``Refining graph structure for
  incomplete multi-view clustering,'' \emph{IEEE Transactions on Neural
  Networks and Learning Systems}, 2022.

\bibitem{liangICDM}
Y.~Liang, D.~Huang, and C.-D. Wang, ``Consistency meets inconsistency: A
  unified graph learning framework for multi-view clustering,'' in \emph{Proc.
  of IEEE International Conference on Data Mining (ICDM)}, 2019, pp.
  1204--1209.

\bibitem{gao2015multi}
H.~Gao, F.~Nie, X.~Li, and H.~Huang, ``Multi-view subspace clustering,'' in
  \emph{Proc. of IEEE International Conference on Computer Vision (ICC)}, 2015,
  pp. 4238--4246.

\bibitem{zhang2017latent}
C.~Zhang, Q.~Hu, H.~Fu, P.~Zhu, and X.~Cao, ``Latent multi-view subspace
  clustering,'' in \emph{Proc. of IEEE Conference on Computer Vision and
  Pattern Recognition (CVPR)}, 2017, pp. 4279--4287.

\bibitem{vidal2011subspace}
R.~Vidal, ``Subspace clustering,'' \emph{IEEE Signal Processing Magazine},
  vol.~28, no.~2, pp. 52--68, 2011.

\bibitem{zhang2020one}
G.-Y. Zhang, Y.-R. Zhou, X.-Y. He, C.-D. Wang, and D.~Huang, ``One-step kernel
  multi-view subspace clustering,'' \emph{Knowledge-Based Systems}, vol. 189,
  p. 105126, 2020.

\bibitem{tang19_tmm}
C.~Tang, X.~Zhu, X.~Liu, M.~Li, P.~Wang, C.~Zhang, and L.~Wang, ``Learning a
  joint affinity graph for multiview subspace clustering,'' \emph{IEEE
  Transactions on Multimedia}, vol.~21, no.~7, pp. 1724--1736, 2019.

\bibitem{Kang2020}
Z.~Kang, X.~Zhao, C.~Peng, H.~Zhu, J.~T. Zhou, X.~Peng, W.~Chen, and Z.~Xu,
  ``Partition level multiview subspace clustering,'' \emph{Neural Networks},
  vol. 122, pp. 279--288, 2020.

\bibitem{DiMSC}
X.~Cao, C.~Zhang, H.~Fu, S.~Liu, and H.~Zhang, ``Diversity-induced multi-view
  subspace clustering,'' in \emph{Proc. of IEEE Conference on Computer Vision
  and Pattern Recognition (CVPR)}, 2015, pp. 586--594.

\bibitem{ChenManSheng}
M.-S. Chen, L.~Huang, C.-D. Wang, and D.~Huang, ``Multi-view clustering in
  latent embedding space,'' in \emph{Proc. of AAAI conference on artificial
  intelligence}, vol.~34, no.~04, 2020, pp. 3513--3520.

\bibitem{MVSC}
Y.~Li, F.~Nie, H.~Huang, and J.~Huang, ``Large-scale multi-view spectral
  clustering via bipartite graph,'' in \emph{Proc. of AAAI Conference on
  Artificial Intelligence}, 2015.

\bibitem{co-clustering-nie2017}
F.~Nie, X.~Wang, C.~Deng, and H.~Huang, ``Learning a structured optimal
  bipartite graph for co-clustering,'' \emph{Advances in Neural Information
  Processing Systems (NeurIPS)}, vol.~30, 2017.

\bibitem{guo2019anchors}
J.~Guo and J.~Ye, ``Anchors bring ease: An embarrassingly simple approach to
  partial multi-view clustering,'' in \emph{Proc. of AAAI Conference on
  Artificial Intelligence}, vol.~33, no.~01, 2019, pp. 118--125.

\bibitem{BIGMC}
L.~Li and H.~He, ``Bipartite graph based multi-view clustering,'' \emph{IEEE
  Transactions on Knowledge and Data Engineering}, 2020.

\bibitem{SFMC}
X.~Li, H.~Zhang, R.~Wang, and F.~Nie, ``Multiview clustering: A scalable and
  parameter-free bipartite graph fusion method,'' \emph{IEEE Transactions on
  Pattern Analysis and Machine Intelligence}, vol.~44, no.~1, pp. 330--344,
  2020.

\bibitem{MSGL}
Z.~Kang, Z.~Lin, X.~Zhu, and W.~Xu, ``Structured graph learning for scalable
  subspace clustering: From single view to multiview,'' \emph{IEEE Transactions
  on Cybernetics}, vol.~52, no.~9, pp. 8976--8986, 2022.

\bibitem{nie2021learning}
F.~Nie, W.~Chang, R.~Wang, and X.~Li, ``Learning an optimal bipartite graph for
  subspace clustering via constrained laplacian rank,'' \emph{IEEE Transactions
  on Cybernetics}, 2021.

\bibitem{SMVSC}
M.~Sun, P.~Zhang, S.~Wang, S.~Zhou, W.~Tu, X.~Liu, E.~Zhu, and C.~Wang,
  ``Scalable multi-view subspace clustering with unified anchors,'' in
  \emph{Proc. of ACM International Conference on Multimedia (ACM MM)}, 2021,
  pp. 3528--3536.

\bibitem{FPMVS-CAG}
S.~Wang, X.~Liu, X.~Zhu, P.~Zhang, Y.~Zhang, F.~Gao, and E.~Zhu, ``Fast
  parameter-free multi-view subspace clustering with consensus anchor
  guidance,'' \emph{IEEE Transactions on Image Processing}, vol.~31, pp.
  556--568, 2021.

\bibitem{LMVSC}
Z.~Kang, W.~Zhou, Z.~Zhao, J.~Shao, M.~Han, and Z.~Xu, ``Large-scale multi-view
  subspace clustering in linear time,'' in \emph{Proc. of AAAI conference on
  Artificial Intelligence}, vol.~34, no.~04, 2020, pp. 4412--4419.

\bibitem{SDAGF}
X.~Lu and S.~Feng, ``Structure diversity-induced anchor graph fusion for
  multi-view clustering,'' \emph{ACM Transactions on Knowledge Discovery from
  Data}, 2022.

\bibitem{huang2019ultra}
D.~Huang, C.-D. Wang, J.-S. Wu, J.-H. Lai, and C.-K. Kwoh, ``Ultra-scalable
  spectral clustering and ensemble clustering,'' \emph{IEEE Transactions on
  Knowledge and Data Engineering}, vol.~32, no.~6, pp. 1212--1226, 2019.

\bibitem{Huang2023}
D.~Huang, C.-D. Wang, and J.-H. Lai, ``Fast multi-view clustering via
  ensembles: Towards scalability, superiority, and simplicity,'' \emph{IEEE
  Transactions on Knowledge and Data Engineering, in press}, 2023.

\bibitem{MLAN}
F.~Nie, G.~Cai, and X.~Li, ``Multi-view clustering and semi-supervised
  classification with adaptive neighbours,'' in \emph{Proc. of AAAI Conference
  on Artificial Intelligence}, 2017.

\bibitem{MVCTM}
S.~Huang, I.~Tsang, Z.~Xu, J.~Lv, and Q.-H. Liu, ``Multi-view clustering on
  topological manifold,'' in \emph{Proceedings of the AAAI Conference on
  Artificial Intelligence}, vol.~36, no.~6, 2022, pp. 6944--6951.

\bibitem{liu2010large}
W.~Liu, J.~He, and S.-F. Chang, ``Large graph construction for scalable
  semi-supervised learning,'' in \emph{Proc. of International Conference on
  Machine Learning (ICML)}, 2010.

\bibitem{von2007tutorial}
U.~Von~Luxburg, ``A tutorial on spectral clustering,'' \emph{Statistics and
  Computing}, vol.~17, no.~4, pp. 395--416, 2007.

\bibitem{fan1949theorem}
K.~Fan, ``On a theorem of weyl concerning eigenvalues of linear transformations
  i,'' \emph{Proceedings of the National Academy of Sciences of the United
  States of America}, vol.~35, no.~11, p. 652, 1949.

\bibitem{Anew}
J.~Huang, F.~Nie, and H.~Huang, ``A new simplex sparse learning model to
  measure data similarity for clustering,'' in \emph{Proc. of International
  Joint Conference on Artificial Intelligence (IJCAI)}, 2015.

\bibitem{ALM}
D.~P. Bertsekas, ``Constrained optimization and lagrange multiplier methods,''
  \emph{Computer Science and Applied Mathematics}, 1982.

\bibitem{Cai2023}
X.~Cai, D.~Huang, G.-Y. Zhang, and C.-D. Wang, ``Seeking commonness and
  inconsistencies: A jointly smoothed approach to multi-view subspace
  clustering,'' \emph{Information Fusion}, vol.~91, pp. 364--375, 2023.

\bibitem{D-Out-Scene}
Z.~Hu, F.~Nie, R.~Wang, and X.~Li, ``Multi-view spectral clustering via
  integrating nonnegative embedding and spectral embedding,'' \emph{Information
  Fusion}, vol.~55, pp. 251--259, 2020.

\bibitem{D-Citeseer}
X.~Liu, L.~Liu, Q.~Liao, S.~Wang, Y.~Zhang, W.~Tu, C.~Tang, J.~Liu, and E.~Zhu,
  ``One pass late fusion multi-view clustering,'' in \emph{Proc. of
  International Conference on Machine Learning (ICML)}, 2021, pp. 6850--6859.

\bibitem{Huang2021}
D.~Huang, C.-D. Wang, H.~{Peng}, J.-H. {Lai}, and C.-K. {Kwoh}, ``Enhanced
  ensemble clustering via fast propagation of cluster-wise similarities,''
  \emph{IEEE Transactions on Systems, Man, and Cybernetics: Systems}, vol.~51,
  no.~1, pp. 508--520, 2021.

\bibitem{Deng23_SACC}
X.~Deng, D.~Huang, D.-H. Chen, C.-D. Wang, and J.-H. Lai, ``Strongly augmented
  contrastive clustering,'' \emph{Pattern Recognition}, vol. 139, p. 109470,
  2023.

\bibitem{EOMSC-CA}
S.~Liu, S.~Wang, P.~Zhang, K.~Xu, X.~Liu, C.~Zhang, and F.~Gao, ``Efficient
  one-pass multi-view subspace clustering with consensus anchors,'' in
  \emph{Proceedings of the AAAI Conference on Artificial Intelligence},
  vol.~36, no.~7, 2022, pp. 7576--7584.

\end{thebibliography}

\begin{IEEEbiography}[{\includegraphics[width=1in,height=1.25in,clip,keepaspectratio]{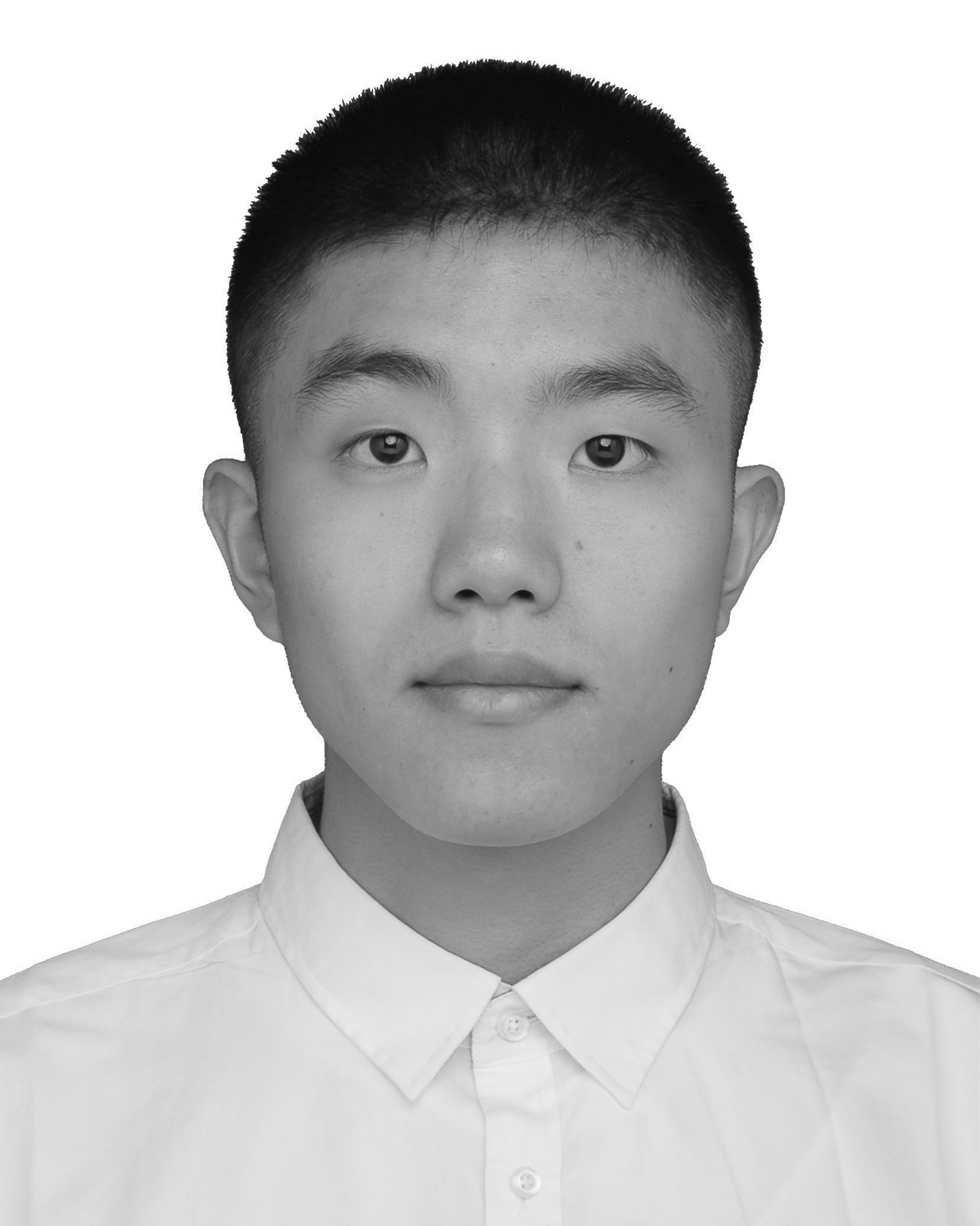}}]{Siguo Fang}
	received the B.S. degree in information and computing sciences from the China Jiliang University, Hangzhou, China, in 2021. He is currently pursuing the master degree in computer science with the College of Mathematics and Informatics, South China Agricultural University, Guangzhou, China. His research interests include multi-view clustering and large-scale clustering.
\end{IEEEbiography}

\begin{IEEEbiography}[{\includegraphics[width=1in,height=1.25in,clip,keepaspectratio]{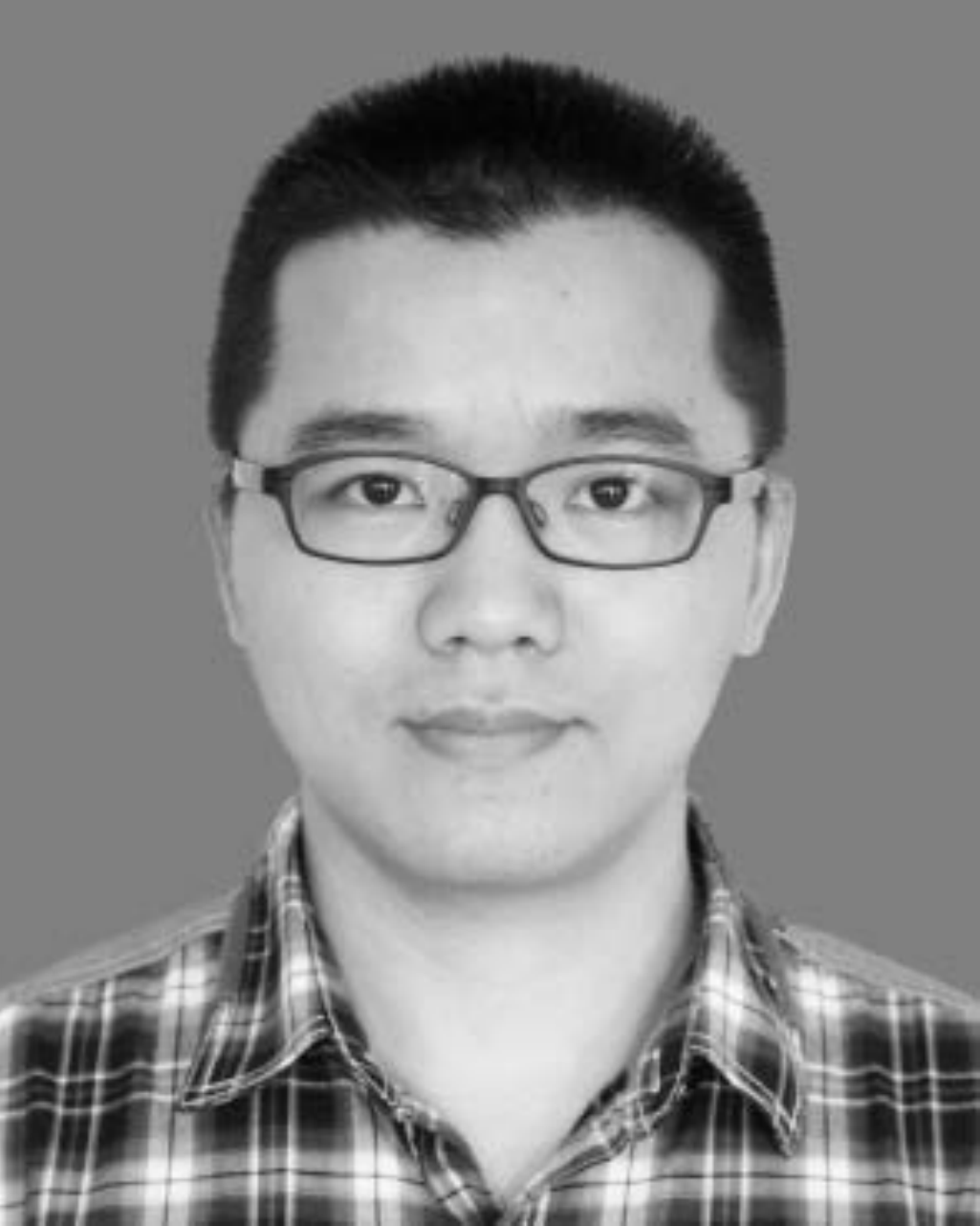}}]{Dong Huang}
	received the B.S. degree in computer science in 2009 from South China University of Technology, Guangzhou, China. He received the M.Sc. degree in computer science in 2011 and the Ph.D. degree in computer science in 2015, both from Sun Yat-sen University, Guangzhou, China. He joined South China Agricultural University in 2015, where he is currently an Associate Professor with the College of Mathematics and Informatics. From July 2017 to July 2018, he was a visiting fellow with the School of Computer Science and Engineering, Nanyang Technological University, Singapore. His research interests include data mining and machine learning. He has published more than 70 papers in international journals and conferences, such as IEEE TKDE, IEEE TNNLS, IEEE TCYB, IEEE TSMC-S, ACM TKDD, SIGKDD, AAAI, and ICDM. He was the recipient of the 2020 ACM Guangzhou Rising Star Award.
\end{IEEEbiography}

\begin{IEEEbiography}[{\includegraphics[width=1in,height=1.25in,clip,keepaspectratio]{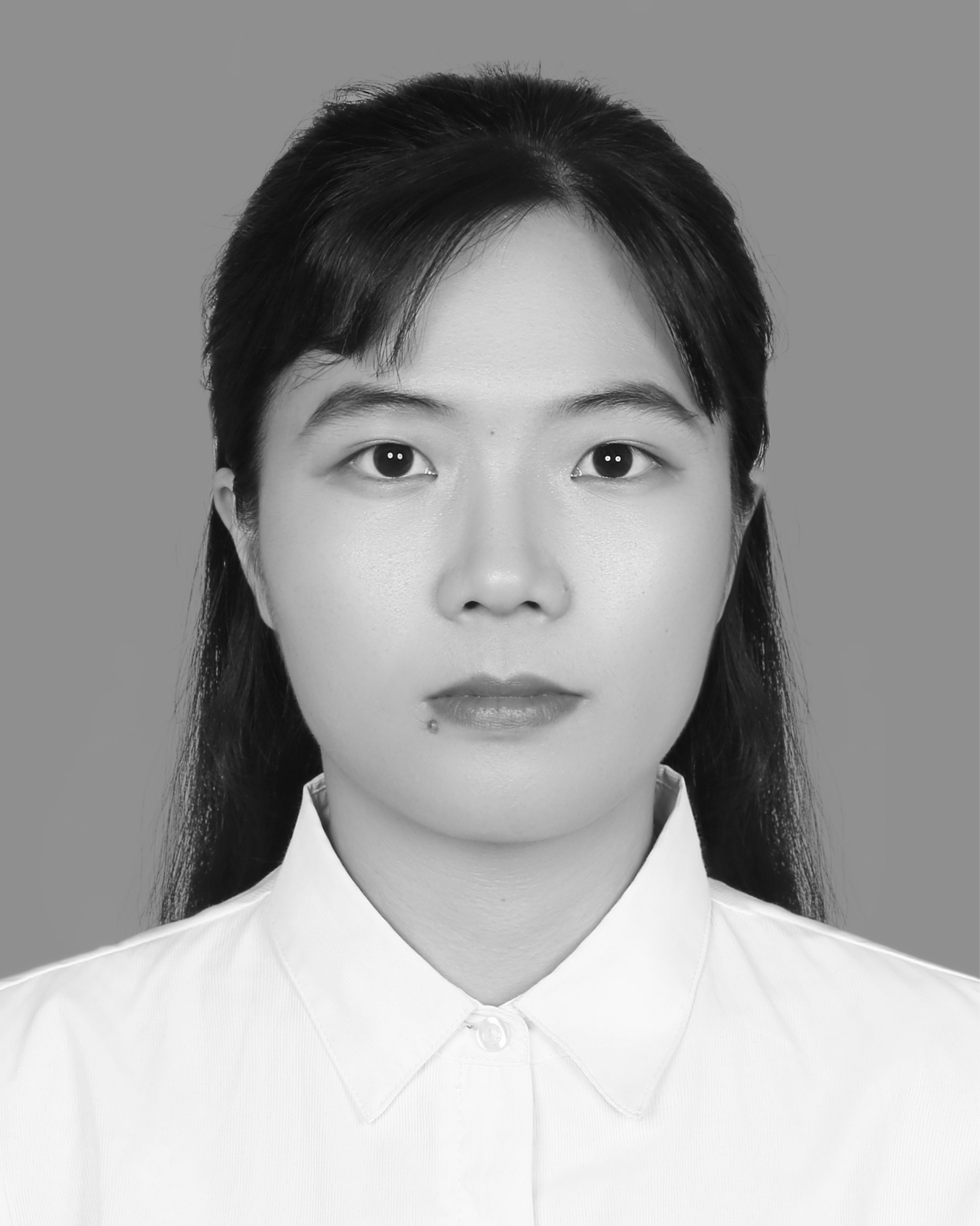}}]{Xiao-Sha Cai}
	received her B.S. degree in software engineering in 2019 and her M.Sc. degree in computer science in 2022, both from South China Agricultural University, Guangzhou, China. She is currently pursuing her Ph.D degree at Sun Yat-sen University, Guangzhou, China. Her research interests include data clustering and information fusion.
\end{IEEEbiography}

\begin{IEEEbiography}[{\includegraphics[width=1in,height=1.25in,clip,keepaspectratio]{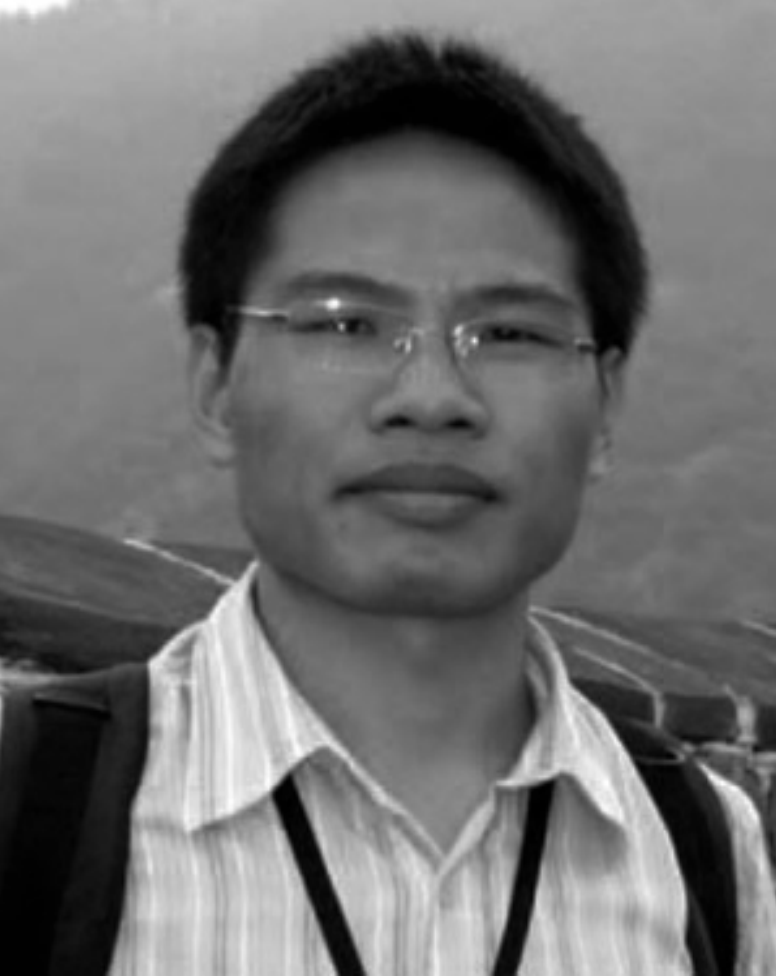}}]{Chang-Dong Wang}
	received the B.S. degree in applied mathematics in 2008, the M.Sc. degree in computer science in 2010, and the Ph.D. degree in computer science in 2013, all from Sun Yat-sen University, Guangzhou, China. He was a visiting student at the University of Illinois at Chicago from January 2012 to November 2012. He is currently an Associate Professor with the School of Data and Computer Science, Sun Yat-sen University, Guangzhou, China. His current research interests include machine learning and data mining. He has published more than 100 scientific papers in international journals and conferences such as IEEE TPAMI, IEEE TKDE, IEEE TNNLS, IEEE TSMC-C, ACM TKDD, Pattern Recognition, SIGKDD, ICDM and SDM. His ICDM 2010 paper won the Honorable Mention for Best Research Paper Award. He was awarded 2015 Chinese Association for Artificial Intelligence (CAAI) Outstanding Dissertation.
\end{IEEEbiography}

\begin{IEEEbiography}[{\includegraphics[width=1in,height=1.25in,clip,keepaspectratio]{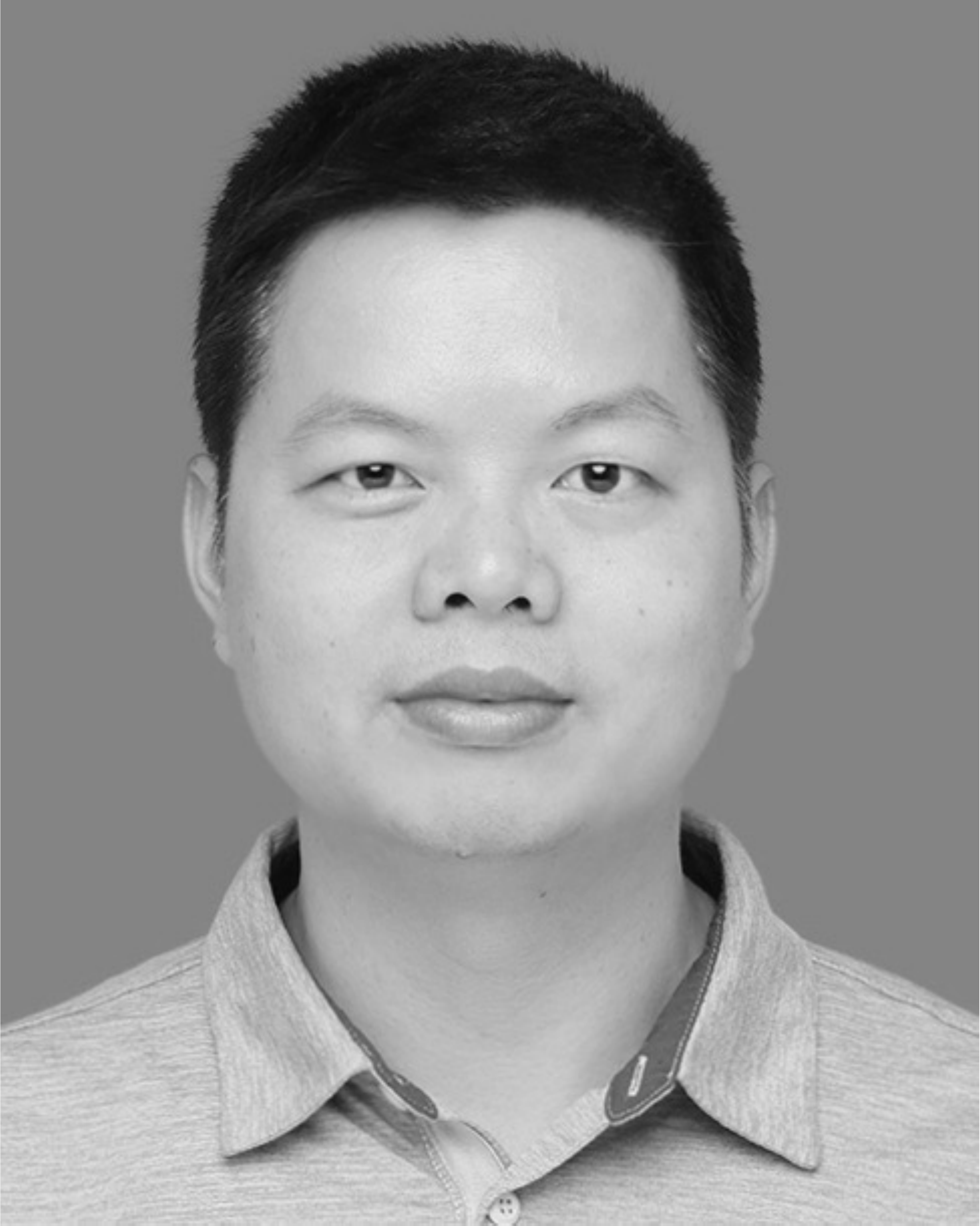}}]{Chaobo He}
	received his Ph.D., M.S., and B.S. degrees from South China Normal University, Guangzhou, China, in 2014, 2007, and 2004, respectively. He is currently a Professor with the School of Computer Science, South China Normal University. His research interests are data mining and social computing. He has published over 30 papers in international journals and conferences.
\end{IEEEbiography}

\begin{IEEEbiography}[{\includegraphics[width=1in,height=1.25in,clip,keepaspectratio]{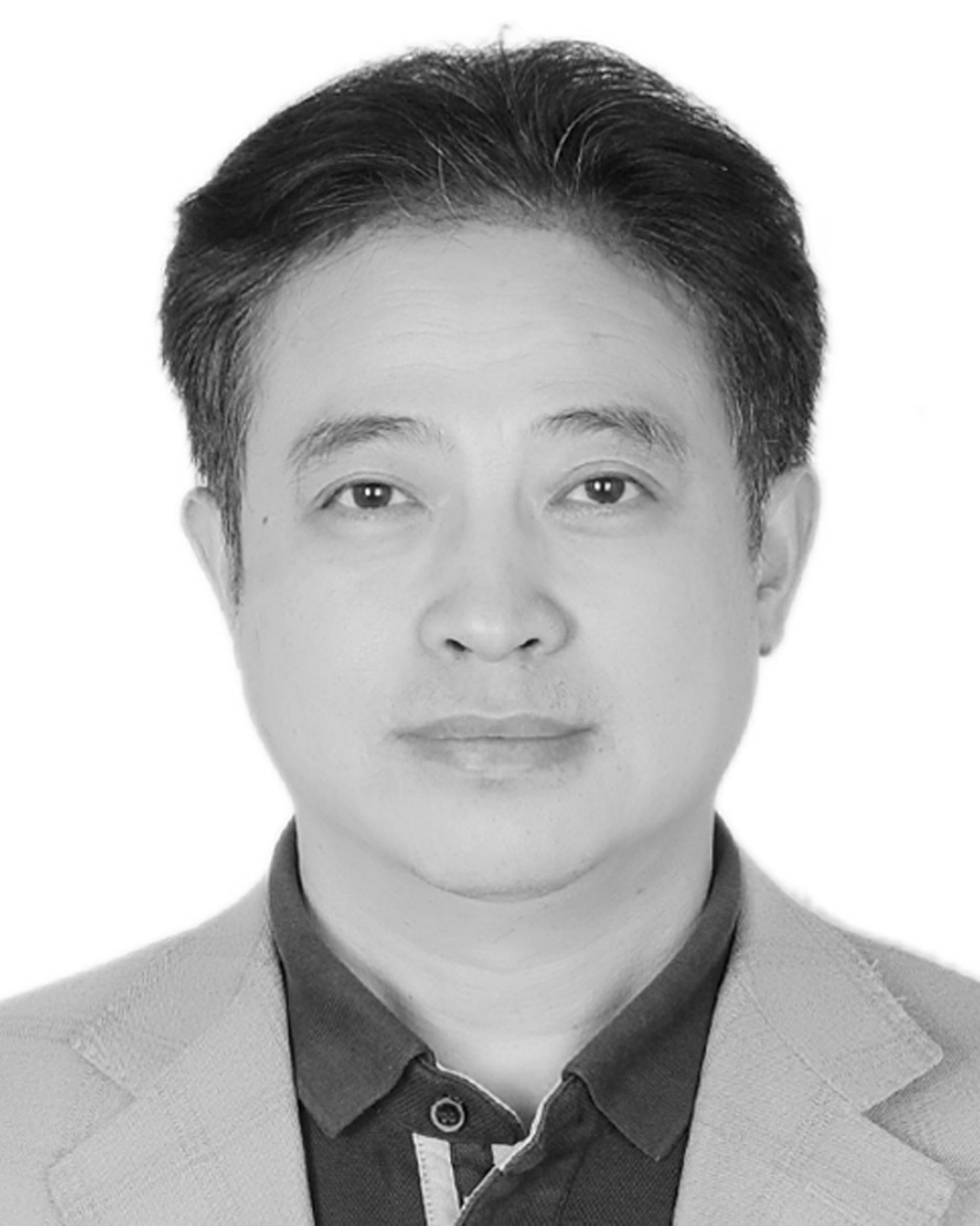}}]{Yong Tang} is the founder of SCHOLAT, a kind of scholar social network. He is now a Professor and Dean of School of Computer Science at South China Normal University. He got his BS and MSc degrees from Wuhan University in 1985 and 1990 respectively, and PhD degree from University of Science and Technology of China in 2001, all in computer science. Before joining South China Normal University in 2009, he was vice Dean of School of Information of Science and Technology at Sun Yat-Sen University. He has published more than 200 papers and books. He has supervised more than 40 PhD students since 2003 and more than 100 Master students since 1996. His main research areas include data and knowledge engineering, social networking and collaborative computing. He currently serves as the director of technical committee on collaborative computing of China Computer Federation (CCF) and the executive vice president of Guangdong Computer Academy. For more information, please visit \url{https://scholat.com/ytang}.
\end{IEEEbiography}

\end{document}